\documentclass[oribib]{llncs}
\usepackage{lipsum}  
\usepackage{makeidx}  
\usepackage{tensor}
\usepackage{amsfonts}
\usepackage{amsmath}
\usepackage[labelformat=simple]{subcaption}
\usepackage{dsfont}
\usepackage{tikz}
\usepackage[hidelinks]{hyperref}
\usepackage{pgfplots}
\usepackage{graphicx} 
\usepackage{siunitx}
\usepackage{placeins}
\usepackage{gensymb}
\usepackage{mathtools}
\usepackage{booktabs}
\usepackage{array}
\usepackage{makecell}
\usepackage[export]{adjustbox}

\newcolumntype{L}[1]{>{\raggedright\let\newline\\\arraybackslash\hspace{0pt}}m{#1}}
\newcolumntype{C}[1]{>{\centering\let\newline\\\arraybackslash\hspace{0pt}}m{#1}}
\newcolumntype{R}[1]{>{\raggedleft\let\newline\\\arraybackslash\hspace{0pt}}m{#1}}

\usepgfplotslibrary{units}

\pgfplotsset{compat=1.16} 

\pgfplotsset{every tick label/.append style={font=\scriptsize}}

\graphicspath{ {images/} }

\definecolor{crimson2143940}{RGB}{214,39,40}
\definecolor{darkgray176}{RGB}{176,176,176}
\definecolor{darkorange25512714}{RGB}{255,127,14}
\definecolor{forestgreen4416044}{RGB}{44,160,44}
\definecolor{lightgray204}{RGB}{204,204,204}
\definecolor{steelblue31119180}{RGB}{31,119,180}

\makeatletter
\renewcommand\subsubsection{\@startsection{subsubsection}{3}{\z@}%
                       {-18\p@ \@plus -4\p@ \@minus -4\p@}%
                       {4\p@ \@plus 2\p@ \@minus 2\p@}%
                       {\normalfont\normalsize\bfseries\boldmath
                        \rightskip=\z@ \@plus 8em\pretolerance=10000 }}
\renewcommand\paragraph{\@startsection{paragraph}{4}{\z@}%
                       {-12\p@ \@plus -4\p@ \@minus -4\p@}%
                       {-0.5em \@plus -0.22em \@minus -0.1em}%
                       {\normalfont\normalsize\bfseries\boldmath}}
\makeatother

\begin{document}
\frontmatter          
\pagestyle{headings}  
\addtocmark{Hamiltonian Mechanics} 
\mainmatter              
\title{On the Evaluation of RGB-D-based\\ Categorical Pose and Shape Estimation}
\titlerunning{Evaluation of RGB-D-based Categorical Pose and Shape Estimation}  
%
\author{Leonard Bruns \and Patric Jensfelt }
\authorrunning{Leonard Bruns et al.} 
%
%
\institute{KTH Royal Institute of Technology,\\Division of Robotics, Perception and Learning, Sweden\\
\email{\{leonardb,patric\}@kth.se}}

\maketitle              

\begin{abstract}
Recently, various methods for 6D pose and shape estimation of objects have been proposed. Typically, these methods evaluate their pose estimation in terms of average precision, and reconstruction quality with chamfer distance. In this work we take a critical look at this predominant evaluation protocol including metrics and datasets. We propose a new set of metrics, contribute new annotations for the Redwood dataset and evaluate state-of-the-art methods in a fair comparison. We find that existing methods do not generalize well to unconstrained orientations, and are actually heavily biased towards objects being upright. We contribute an easy-to-use evaluation toolbox with well-defined metrics, method and dataset interfaces, which readily allows evaluation and comparison with various state-of-the-art approaches (\url{https://github.com/roym899/pose_and_shape_evaluation}).
\end{abstract}

\section{Introduction}

We consider the problem of pose and shape estimation on a per-category level. Classic grasp and motion planning methods often assume full knowledge of pose and shape to be available, which makes them difficult to apply with partial sensor information. Estimating the full shape and pose promises to bridge this gap from partial sensor information to an actionable representation. While shape reconstruction by itself is sufficient for some tasks, categorical pose estimation additionally provides a reference frame. This categorical reference frame could, for example, further enable aligning objects, and pose dependent grasp computation (e.g., an upside down mug has to be grasped differently from an upright mug).

Over the last two years various learning-based methods for categorical pose and shape estimation have been introduced. Most methods build and evaluate on the two datasets proposed by \cite{wang2019normalized}. The CAMERA dataset is a large dataset of real RGB-D tabletop scenes with synthetic objects generated on top of the table. The REAL dataset is a smaller real-world dataset of tabletop sequences with objects, that have been scanned and tracked for the purpose of evaluating categorical pose estimation. Notably, both datasets only contain upright objects, which opens the question how well existing methods generalize to less constrained settings.

To answer this question of how well existing methods generalize to less constrained settings, we contribute a set of annotations to evaluate unconstrained 6D pose and shape estimation. Our annotations consist of meshes and poses for handheld objects of three categories in the Redwood dataset \cite{choi2016large}. In that dataset the objects are freely rotated in front of the camera, and the orientations vary significantly more than those included in the datasets by \cite{wang2019normalized}.

Most methods evaluate pose estimation by following the same evaluation protocol as initially proposed by \cite{wang2019normalized}. However, the method in~\cite{wang2019normalized} combines mask detection and pose estimation into a single network and therefore evaluate pose estimation with \emph{average precision} (AP), which is a common detection metric. Many of the subsequent methods however assume the mask to be an input to their method, which makes AP an unnatural evaluation metric as it makes the results unnecessarily difficult to interpret. Therefore, we propose a set of simpler metrics, that use ground truth masks and classes to evaluate pose estimation.

Pose estimation with shape reconstruction has first been demonstrated by \cite{chen2020learning} and \cite{tian2020shape}. Both methods independently used chamfer distance as the reconstruction metric, which is now commonly used by methods performing shape reconstruction. However, \cite{tatarchenko2019single} showed that chamfer distance is \emph{not} a good measure of reconstruction quality and other metrics better correlate with perceived reconstruction quality. Therefore, with this work we advocate for a new set of reconstruction metrics and propose a new evaluation protocol for both shape reconstruction and pose estimation.

To summarize, our contributions are:
\begin{itemize}
    \item a well-defined evaluation protocol,
    \item a challenging set of novel annotations to evaluate unconstrained pose and shape estimation,
    \item a fair evaluation of various state-of-the-art methods, and
    \item an open source evaluation toolbox for the task of categorical pose and shape estimation.
\end{itemize}

\section{Related Work}


Since the introduction of the BOP benchmark suite \cite{hodavn2020bop}, a lot of progress has been made in the task of instance-level pose estimation, where a mesh of the target object is available. While such pose estimation with known meshes has made remarkable progress, the more general task of category-level pose estimation has only recently received more attention. Compared to instance-level pose estimation categorical pose and shape estimation is more difficult due to the large possible variations in shape and appearance even for a single category.



Wang et al.\ \cite{wang2019normalized} introduced the first deep learning-based method to address the 6D pose estimation problem on a per-category level. They introduced two datasets: a synthetic dataset that combines real scenes with meshes from the ShapeNet dataset \cite{chang2015} and a smaller real-world dataset that is mainly used for finetuning and evaluation. Their method is based on \emph{normalized object coordinate space} (NOCS), in which objects of one category have a common alignment. The projection of the NOCS coordinates in the image plane (also called NOCS map) is predicted by extending Mask R-CNN \cite{he2017mask} with an additional head. From this prediction, the 6D pose and scale can be estimated by employing Umeyama algorithm \cite{umeyama1991least} with RANSAC \cite{fischler1981random} for outlier removal.

The NOCS-map was predicted using only the RGB information. Since geometry typically varies less than appearance for a fixed category, several methods were proposed to more directly incorporate the observed point set in the prediction. Chen et al.\ \cite{chen2020learning} introduced \emph{canonical shape space} (CASS) which regresses orientation and position directly from the cropped image and observed point set. As a byproduct of their method they also reconstruct the full canonical point set. Tian et al.\ \cite{tian2020shape} introduced \emph{shape prior deformation} (SPD), which uses a canonical point set and predicts a deformation based on the observed RGB-D information. CR-Net \cite{wang2021category} and SGPA \cite{chen2021sgpa} are two extensions to the original idea of SPD. CR-Net uses a recurrent architecture to iteratively deform the canonical point set, and SGPA uses a transformer architecture to more effectively adjust the canonical point set. Recently, good results were demonstrated by only training on synthetically generated views of ShapeNet meshes \cite{akizuki2021}.

Other methods such as \cite{chen2021fs} and \cite{lin2021dualposenet} predict pose and bounding box without reconstructing the full shape of the object. For the evaluation presented in this work we limit ourselves to methods performing both reconstruction and pose estimation, although our evaluation protocol could in principle be used for pure pose estimation methods as well.

Aside from these RGB-D-based methods, more RGB-based methods have been proposed. Chen et al.\ \cite{chen2020category} proposed an analysis-by-synthesis framework in which the latent representation of a generative model is iteratively optimized to fit the observed color image. The generative model allows to generate novel views of the object, but a full reconstruction is not readily available. Lee et al.\ \cite{lee2021category} introduced a framework for estimating a mesh from an RGB image, while Engelmann et al.\ \cite{engelmann2021points} proposes to reconstruct shapes in a representation agnostic way by classifying the closest matching object from a database.


\paragraph{Evaluation} The most established benchmark dataset for categorical pose estimation is the REAL275 dataset proposed by \cite{wang2019normalized}. We will take a critical look at that dataset in Section \ref{section:real275} and show that it only evaluates a constrained set of orientations, hiding inherent difficulties of the task such as multimodal orientation distributions. \cite{wang2019normalized} also proposed \emph{average precision} as a metric to evaluate pose estimation.

To evaluate shape reconstruction most papers currently use chamfer distance (CD) \cite{akizuki2021,chen2020learning,tian2020shape}, which was introduced to measure the difference of point sets by \cite{fan2017point}. However, \cite{tatarchenko2019single} noted that CD is not robust to outliers, that is, the distance of outliers affects the metric. The authors therefore advocate to use a robust, thresholded metric such as F-score to measure reconstruction quality \cite{Knapitsch2017}.


\section{Evaluation Protocol}

In this section we will discuss the existing and proposed evaluation protocol. We will start by formally defining the problem of categorical pose and shape estimation. We will then discuss metrics to evaluate and compare different solutions to this problem. Finally, we discuss the evaluation datasets.

\subsection{Problem Definition}\label{section:definition}

Let $\mathbf{I}\in\mathbb{R}^{H\times W\times 3}$ be an RGB image, $\mathbf{D}\in\mathbb{R}^{H\times W}$ be a depth map and $\mathbf{P}\in\mathbb{R}^{3\times 4}$ be the projection matrix of the associated camera. Further, let $\tensor[^i]{\mathbf{T}}{_j}$ be the homogeneous transformation matrix, that transforms a point $\tensor[^j]{\mathbf{p}}{}$ from frame $j$ to frame $i$, that is, $\tensor[^i]{\mathbf{p}}{}=\tensor[^i]{\mathbf{T}}{_j}\tensor[^j]{\mathbf{p}}{}$. Note that depending on the context, $\tensor[^i]{\mathbf{T}}{_j}$ can also be interpreted as the 6D pose of frame $j$ in frame $i$. Let $\tensor[^i]{\mathbf{R}}{_j}$ and $\tensor[^i]{\mathbf{t}}{_j}$ further denote the rotation matrix and translation vector, out of which $\tensor[^i]{\mathbf{T}}{_j}$ is composed.

We will use $\mathcal{O}$ to denote a 3D object, and $\mathcal{B}(\mathcal{O})$ to denote the axis aligned bounding box of $\mathcal{O}$ in $\mathcal{O}$'s frame $\mathrm{o}$. We will assume that the center of $\mathcal{B}(\mathcal{O})$ is at the origin of frame $\mathrm{o}$, and that the camera is at the origin of frame $\mathrm{w}$ (world). We further assume that transforms can be applied to 3D objects and bounding boxes, for example, $\tensor[^{\mathrm{w}}]{\mathcal{O}}{}=\tensor[^{\mathrm{w}}]{\mathbf{T}}{_{\mathrm{o}}}\mathcal{O}$. Following this notation, note that there is a difference between $\mathcal{B}({\tensor[^{\mathrm{w}}]{\mathcal{O}}{}})$ and $\tensor[^{\mathrm{w}}]{\mathbf{T}}{_{\mathrm{o}}}\mathcal{B}(\mathcal{O})$. The first one is an \emph{axis-aligned bounding box} (AABB), the second is an \emph{oriented bounding box} (OBB).

\begin{problem}{(Categorical Pose and Shape Estimation)}\label{problem}
    Given $(\mathbf{I},\mathbf{D},\mathbf{P})$ imaging an object $\mathcal{O}$ of known category $c$ at pose $\tensor[^{\mathrm{w}}]{\mathbf{T}}{_{\mathrm{o}}}$, and given the mask $\mathbf{M}$ of visible points of the object in the image, find estimates $\smash{\widetilde{\mathcal{O}}}$ and $\tensor[^{\mathrm{w}}]{\widetilde{\mathbf{T}}}{_{\mathrm{o}}}$ of $\mathcal{O}$ and $\tensor[^{\mathrm{w}}]{\mathbf{T}}{_{\mathrm{o}}}$, respectively.
\end{problem}

Similarly one could define the problems of categorical pose estimation (estimate $\tensor[^{\mathrm{w}}]{\mathbf{T}}{_{\mathrm{o}}}$ only) and categorical pose and size estimation (estimate $\tensor[^{\mathrm{w}}]{\mathbf{T}}{_{\mathrm{o}}}$ and $\mathcal{B}(\mathcal{O})$).


\subsection{Metrics}\label{section:metrics}

Various metrics exist to assess how well a method solves Problem \ref{problem}. Currently, the predominant evaluation metric is \emph{average precision} \cite{wang2019normalized,chen2020learning,wang2021category,chen2021sgpa,akizuki2021,lin2021dualposenet,chen2021fs,lee2021category} for pose estimation, and \emph{chamfer distance} \cite{fan2017point,chen2020learning,wang2021category,chen2021sgpa,akizuki2021,lee2021category} for shape reconstruction. In this section we will introduce these metrics and discuss several issues with them. Afterwards we will advocate for \emph{precision} (contrary to average precision) and \emph{f-score} to evaluate pose estimation and shape reconstruction, respectively.

We first define similarity functions for transforms and objects. These are later used to define the evaluation metrics for Problem \ref{problem}.
\begin{definition}
    Let $d(\tensor[^{\mathrm{w}}]{\mathbf{T}}{_{\mathrm{o}}}, \tensor[^{\mathrm{w}}]{\widetilde{\mathbf{T}}}{_{\mathrm{o}}})$ denote the translation error between the ground truth transform and estimated transform, that is,
    \begin{equation}
        d(\tensor[^{\mathrm{w}}]{\mathbf{T}}{_{\mathrm{o}}},\tensor[^{\mathrm{w}}]{\widetilde{\mathbf{T}}}{_{\mathrm{o}}})=\lVert\tensor[^{\mathrm{w}}]{\mathbf{t}}{_{\mathrm{o}}} - \tensor[^{\mathrm{w}}]{\tilde{\mathbf{t}}}{_{\mathrm{o}}}\rVert_2.
    \end{equation}
\end{definition}

\begin{definition}
    Let $\delta(\tensor[^{\mathrm{w}}]{\mathbf{T}}{_{\mathrm{o}}}, \tensor[^{\mathrm{w}}]{\widetilde{\mathbf{T}}}{_{\mathrm{o}}})$ denote the rotation error between the ground truth transform and estimated transform, that is,
    \begin{equation}
        \delta(\tensor[^{\mathrm{w}}]{\mathbf{T}}{_{\mathrm{o}}},\tensor[^{\mathrm{w}}]{\widetilde{\mathbf{T}}}{_{\mathrm{o}}})=\left|\frac{\mathrm{trace}(\tensor[^{\mathrm{w}}]{\mathbf{R}}{_{\mathrm{o}}}\tensor*[^{\mathrm{w}}]{\widetilde{\mathbf{R}}}{_{\mathrm{o}}^{-1}})-1}{2}\right|.
    \end{equation}
\end{definition}

\begin{definition}
    Let $\mathrm{IoU}(\mathcal{B}_1,\mathcal{B}_2)$ denote the true intersection over union (IoU) of two tight, oriented bounding boxes \cite{ahmadyan2021objectron}. Further, let the axis aligned IoU between two objects be defined by
    \begin{equation}
        \mathrm{IoU}^+(\tensor[^{\mathrm{w}}]{\mathcal{O}}{},\tensor[^{\mathrm{w}}]{\widetilde{\mathcal{O}}}{})=\mathrm{IoU}\left(\mathcal{B}({\tensor[^{\mathrm{w}}]{\mathcal{O}}{}}),\mathcal{B}({\tensor[^{\mathrm{w}}]{\widetilde{\mathcal{O}}}{}})\right),
    \end{equation}
    and the true IoU using tight, oriented bounding boxes by
    \begin{equation}
        \mathrm{IoU}(\tensor[^{\mathrm{w}}]{\mathcal{O}}{},\tensor[^{\mathrm{w}}]{\widetilde{\mathcal{O}}}{})=\mathrm{IoU}\left(\tensor[^{\mathrm{w}}]{\mathbf{T}}{_{\mathrm{o}}}\mathcal{B}(\mathcal{O}),\tensor[^{\mathrm{w}}]{\widetilde{\mathbf{T}}}{_{\mathrm{o}}}\mathcal{B}(\widetilde{\mathcal{O}})\right).
    \end{equation}
\end{definition}
The current evaluation protocol \cite{wang2019normalized} uses $\mathrm{IoU}^+$ instead of $\mathrm{IoU}$, despite being less accurate. Our implementation follows \cite{ahmadyan2021objectron} computing $\mathrm{IoU}$.

\subsubsection{Chamfer distance}\label{section:cd}

\emph{Chamfer distance} (CD) in the context of shape reconstruction was introduced by \cite{fan2017point} to differentiably measure the difference of point sets. 
\begin{definition}
    Let $\mathcal{S}\subset \mathbb{R}^3$ and $\widetilde{\mathcal{S}}\subset\mathbb{R}^3$ denote point sets sampled from the surfaces of $\mathcal{O}$ and $\widetilde{\mathcal{O}}$, respectively. We define CD as
    \begin{equation}
        \mathrm{CD}(\mathcal{S},\widetilde{\mathcal{S}})=\frac{1}{2|\mathcal{S}|}\sum_{\mathbf{x}\in \mathcal{S}}\min_{\mathbf{y}\in\widetilde{\mathcal{S}}}\lVert \mathbf{x} - \mathbf{y} \rVert_2 + \frac{1}{2|\widetilde{\mathcal{S}}|}\sum_{\mathbf{y}\in \widetilde{\mathcal{S}}}\min_{\mathbf{x}\in{\mathcal{S}}}\lVert \mathbf{x} - \mathbf{y} \rVert_2.
    \end{equation}
\end{definition}
It is easiest to interpret as the mean Euclidean distance from a point in one point set to the closest point in the other set. Note that slightly different CD versions exist such as squared versions, and ones using the sum instead of arithmetic mean. 

\begin{figure}
    \centering
    \begin{subfigure}[b]{0.4\textwidth}
        \centering
        \begin{tikzpicture}
            \node (1) at (0,0) {\includegraphics[width=1.8cm]{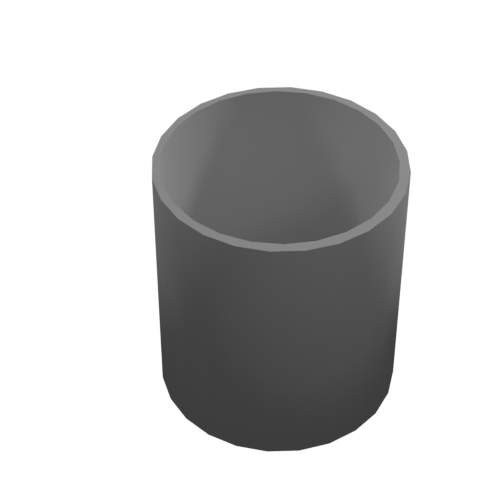}};
            \node (2) at (-1.4,-2) {\includegraphics[width=1.8cm]{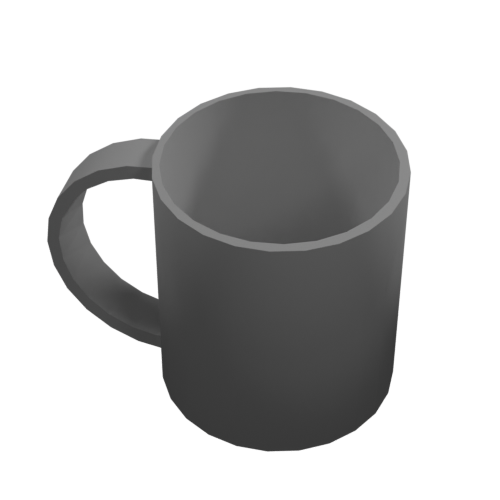}};
            \node (3) at (1.4,-2) {\includegraphics[width=1.8cm]{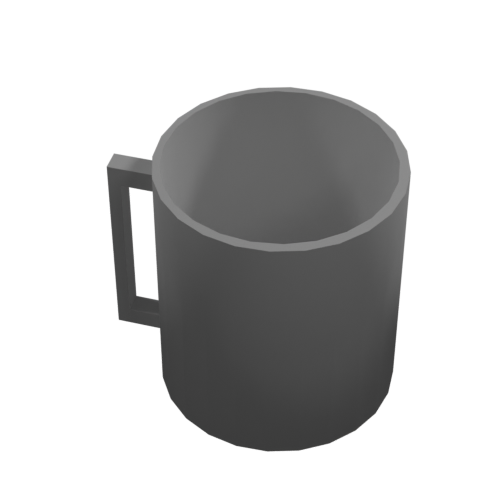}};
            \draw[<->] (-0.3,0.7) to [out=110,in=70,looseness=2]  node[midway, above] {$\mathrm{GT}\leftrightarrow\mathrm{GT}$} (0.5,0.7);
            \draw[<->] (-0.3,-0.7) -- (-0.9,-1.4) node[midway,left] {$\mathrm{GT}\leftrightarrow\mathrm{1}$};
            \draw[<->] (0.5,-0.7) -- (1.1,-1.4) node[midway,right] {$\mathrm{GT}\leftrightarrow\mathrm{2}$};;
        \end{tikzpicture}
        \vspace{0cm}
        \subcaption{Meshes}\label{fig:example_meshes}
    \end{subfigure}
    \begin{subfigure}[b]{0.59\textwidth}
\begin{tikzpicture}

\definecolor{darkgray176}{RGB}{176,176,176}
\definecolor{darkorange25512714}{RGB}{255,127,14}
\definecolor{forestgreen4416044}{RGB}{44,160,44}
\definecolor{steelblue31119180}{RGB}{31,119,180}

\begin{axis}[
width=\textwidth,
height=0.7\textwidth,
log basis x={10},
tick align=outside,
ylabel near ticks,
tick pos=left,
x grid style={darkgray176},
xlabel={\# samples},
xmajorgrids,
xmin=63.0957344480193, xmax=1584893.19246111,
xmode=log,
xtick style={color=black},
y grid style={darkgray176},
ylabel={$\mathrm{CD} / \mathrm{m}$},
ymajorgrids,
ymin=-0.000357484703618492, ymax=0.0103406887394132,
ytick style={color=black}
]
\addplot [very thick, steelblue31119180]
table {%
100 0.00887207639653745
120.679264063933 0.00811490515495286
145.634847750124 0.00737640696402916
175.751062485479 0.00695920815044473
212.095088792019 0.00645543000659581
255.954792269954 0.0059451813218451
308.884359647748 0.00543289692207538
372.759372031494 0.00507654931112869
449.843266896944 0.00468115914739195
542.867543932386 0.00445268615741683
655.128556859551 0.00398067513402519
790.60432109077 0.0037874671856085
954.095476349994 0.00343718972446623
1151.39539932645 0.0031133275135468
1389.49549437314 0.00297498601622347
1676.83293681101 0.00283442254715392
2023.58964772516 0.00256129121685061
2442.05309454865 0.00239621792731292
2947.05170255181 0.00220447533126299
3556.48030622313 0.00203953458271211
4291.93426012878 0.00188323481923328
5179.47467923121 0.00174113533545296
6250.55192527397 0.00161242185762149
7543.12006335462 0.0014613526413073
9102.98177991522 0.00132308859683913
10985.4114198756 0.0012215272464772
13257.1136559011 0.00110455593555721
15998.5871960606 0.00101882455580813
19306.9772888325 0.000921454244140444
23299.5181051537 0.000843655138763264
28117.6869797423 0.000765090979120931
33932.2177189533 0.000697895210917357
40949.1506238042 0.000636828998799428
49417.1336132384 0.000577418015645057
59636.2331659464 0.000527715202747646
71968.5673001151 0.000479635930448023
86851.1373751352 0.000434957964785117
104811.313415469 0.000397714997977767
126485.52168553 0.000361722779532943
152641.796717523 0.000329396955295915
184206.996932672 0.000299906837432821
222299.648252619 0.000273720327817266
268269.579527973 0.000249049251272778
323745.754281764 0.000226618350209181
390693.993705461 0.00020587969481896
471486.636345739 0.000187550025506476
568986.602901829 0.000171009185360244
686648.8450043 0.000155394285924479
828642.772854684 0.000141521151359487
1000000 0.000128795907428403
};
\addplot [very thick, darkorange25512714]
table {%
100 0.0098544081283663
120.679264063933 0.00962652215559299
145.634847750124 0.00871739929352431
175.751062485479 0.00861149963842358
212.095088792019 0.00705867486833976
255.954792269954 0.00668049429555589
308.884359647748 0.00644181601503226
372.759372031494 0.00613072822287869
449.843266896944 0.00548191864150449
542.867543932386 0.00525096216375182
655.128556859551 0.00497317586246658
790.60432109077 0.00460763199779269
954.095476349994 0.00446576338699611
1151.39539932645 0.00416249074950407
1389.49549437314 0.00398510145224972
1676.83293681101 0.00374062820383371
2023.58964772516 0.00355717170772655
2442.05309454865 0.00340009360767452
2947.05170255181 0.00322651918997893
3556.48030622313 0.00301841729856438
4291.93426012878 0.00287739274449161
5179.47467923121 0.00273454560434042
6250.55192527397 0.00260538875676169
7543.12006335462 0.00247338948015978
9102.98177991522 0.00233461605278349
10985.4114198756 0.00223530355250698
13257.1136559011 0.00213103868571683
15998.5871960606 0.00202804833443362
19306.9772888325 0.00194276518539997
23299.5181051537 0.00186040937090073
28117.6869797423 0.00178794026856257
33932.2177189533 0.00172376734160445
40949.1506238042 0.00166123787363394
49417.1336132384 0.00160529434551679
59636.2331659464 0.00155685082883245
71968.5673001151 0.00150897009930137
86851.1373751352 0.00146569545489431
104811.313415469 0.00142658055001089
126485.52168553 0.00139234505520267
152641.796717523 0.00135945238207842
184206.996932672 0.00133038133924684
222299.648252619 0.00130538776142812
268269.579527973 0.00128001706057
323745.754281764 0.00125953347342331
390693.993705461 0.00123845451696843
471486.636345739 0.00122125940061886
568986.602901829 0.00120481098984016
686648.8450043 0.00118910474316047
828642.772854684 0.00117534673574433
1000000 0.00116292540672353
};
\addplot [very thick, forestgreen4416044]
table {%
100 0.00839988869119809
120.679264063933 0.00835949652358131
145.634847750124 0.00835434029461695
175.751062485479 0.00721793222931058
212.095088792019 0.00656990781298667
255.954792269954 0.00617076557924616
308.884359647748 0.0058276374068993
372.759372031494 0.00547335196372112
449.843266896944 0.00492880321600788
542.867543932386 0.00452648536686067
655.128556859551 0.00431980578638745
790.60432109077 0.00397668861503106
954.095476349994 0.00363937041418539
1151.39539932645 0.00348773195750739
1389.49549437314 0.00330799232536578
1676.83293681101 0.00299811414573868
2023.58964772516 0.002870135104484
2442.05309454865 0.00262952435943032
2947.05170255181 0.00245728501768351
3556.48030622313 0.00228008169161143
4291.93426012878 0.00215124842140402
5179.47467923121 0.00200537545972758
6250.55192527397 0.0018698734703572
7543.12006335462 0.00171509240465301
9102.98177991522 0.00159105084932836
10985.4114198756 0.00147416182514654
13257.1136559011 0.00137870346986043
15998.5871960606 0.00127567259806342
19306.9772888325 0.00119081631096961
23299.5181051537 0.00110525153643973
28117.6869797423 0.00103052821670086
33932.2177189533 0.000964132627335628
40949.1506238042 0.000899538792831904
49417.1336132384 0.000843889997326636
59636.2331659464 0.000795266772089992
71968.5673001151 0.000745811491478122
86851.1373751352 0.00070509139105524
104811.313415469 0.000665281446276757
126485.52168553 0.000628973294919481
152641.796717523 0.000597832743348116
184206.996932672 0.000568245935881892
222299.648252619 0.000542283444245131
268269.579527973 0.000517794789822584
323745.754281764 0.000494619492178697
390693.993705461 0.000474928658104126
471486.636345739 0.000456973967035564
568986.602901829 0.000440197083590331
686648.8450043 0.000425171137549512
828642.772854684 0.00041081125649559
1000000 0.000398668872015543
};
\legend{$\mathrm{GT}\leftrightarrow\mathrm{GT}$,$\mathrm{GT}\leftrightarrow\mathrm{1}$,$\mathrm{GT}\leftrightarrow\mathrm{2}$}
\end{axis}

\end{tikzpicture}
        \subcaption{Chamfer distance for varying number of samples}
    \end{subfigure}
    \caption{Visualization of the effect of varying number of samples on the chamfer distance. Note that particularly the relative difference between $\mathrm{CD}_1$ and $\mathrm{CD}_2$ varies significantly. This is because the majority of the error stems from sparse sampling, not from actual differences in geometry. All mugs have been scaled to be \SI{10}{cm} tall.}
    \label{fig:cd}
\end{figure}
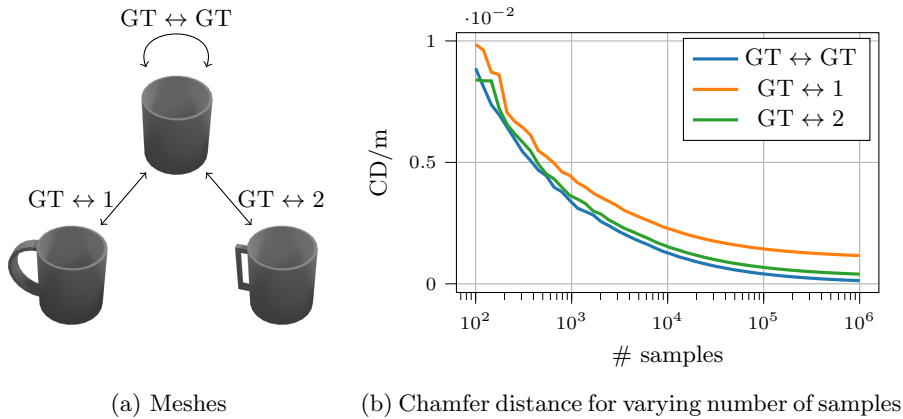

In Fig.\ \ref{fig:cd} we visualize potential issues with CD as an evaluation metric. Consider the two mugs with different handles as reconstructions of the mug without handle. The relative difference between these reconstruction based on CD varies significantly depending on the number of samples. Furthermore, a large number of samples is required for CD to converge.

Notably, the number of ground-truth samples is left unspecified by most methods. This issue gets further amplified due to methods performing reconstruction with varying number of samples as noted by \cite{akizuki2021}. We therefore discourage further use of CD for evaluation purposes.

\subsubsection{Reconstruction F-score}\label{section:fscore}

Following \cite{tatarchenko2019single}, we advocate to use F-score instead of CD to evaluate shape reconstruction. Furthermore, note that we can evaluate shape reconstruction in the object frame, or in the world frame, taking into account $\tensor*[^{\mathrm{w}}]{\mathbf{T}}{_{\mathrm{o}}}$ and $\tensor*[^{\mathrm{w}}]{\widetilde{\mathbf{T}}}{_{\mathrm{o}}}$. While previous methods evaluate in the object frame (i.e., assuming perfect alignment based on canonical reference frame), we believe it is better to evaluate posed reconstruction in the world frame as it correlates more directly with downstream usability of the full estimate.

\begin{definition}
    Let $\mathcal{S}\subset \mathbb{R}^3$ and $\widetilde{\mathcal{S}}\subset\mathbb{R}^3$ denote point sets sampled from the surfaces of $\mathcal{O}$ and $\widetilde{\mathcal{O}}$, respectively.
    Given an application specific threshold $\Delta$, we define reconstruction recall as
    \begin{equation}
        r_\Delta = \frac{1}{|\mathcal{S}|}\sum_{\mathbf{x}\in \mathcal{S}}\left[\min_{\mathbf{y}\in\widetilde{\mathcal{S}}}\lVert \mathbf{x} - \mathbf{y} \rVert_2 < \Delta\right],
    \end{equation}
    and reconstruction precision as
    \begin{equation}
        p_\Delta = \frac{1}{|\widetilde{\mathcal{S}}|}\sum_{\mathbf{y}\in \widetilde{\mathcal{S}}}\left[\min_{\mathbf{x}\in\mathcal{S}}\lVert \mathbf{x} - \mathbf{y} \rVert_2 < \Delta\right],
    \end{equation}
    where $\left[\cdot\right]$ denotes the Iverson bracket. Finally, we define F-score as the harmonic mean of precision and recall
    \begin{equation}
        F_\Delta = \frac{2}{p^{-1}_\Delta+r^{-1}_\Delta}.
    \end{equation}
\end{definition}
Note that $\Delta$ should be adjusted depending on the application and sensor. For the tabletop items contained in the datasets we propose to use $\Delta=\SI{1}{cm}$.

\begin{figure}
    \centering
    \begin{subfigure}[b]{0.59\textwidth}
\begin{tikzpicture}

\definecolor{darkgray176}{RGB}{176,176,176}
\definecolor{darkorange25512714}{RGB}{255,127,14}
\definecolor{forestgreen4416044}{RGB}{44,160,44}
\definecolor{steelblue31119180}{RGB}{31,119,180}

\begin{axis}[
width=\textwidth,
height=0.7\textwidth,
log basis x={10},
tick align=outside,
tick pos=left,
x grid style={darkgray176},
legend pos=south east,
xlabel={\# samples},
xmajorgrids,
xmin=63.0957344480193, xmax=1584893.19246111,
xmode=log,
xtick style={color=black},
y grid style={darkgray176},
ylabel={$F_{1\mathrm{cm}}$},
ymajorgrids,
ymin=0.539486725663717, ymax=1.02192920353982,
ytick style={color=black}
]
\addplot [very thick, steelblue31119180]
table {%
100 0.693237410071943
120.679264063933 0.662264150943396
145.634847750124 0.781528178641957
175.751062485479 0.854046822742475
212.095088792019 0.860687516154045
255.954792269954 0.895559274037843
308.884359647748 0.952454700325228
372.759372031494 0.981175430844012
449.843266896944 0.988864142538975
542.867543932386 0.99446152160965
655.128556859551 0.997709339609969
790.60432109077 1
954.095476349994 0.998950682056663
1151.39539932645 1
1389.49549437314 1
1676.83293681101 1
2023.58964772516 1
2442.05309454865 1
2947.05170255181 1
3556.48030622313 1
4291.93426012878 1
5179.47467923121 1
6250.55192527397 1
7543.12006335462 1
9102.98177991522 1
10985.4114198756 1
13257.1136559011 1
15998.5871960606 1
19306.9772888325 1
23299.5181051537 1
28117.6869797423 1
33932.2177189533 1
40949.1506238042 1
49417.1336132384 1
59636.2331659464 1
71968.5673001151 1
86851.1373751352 1
104811.313415469 1
126485.52168553 1
152641.796717523 1
184206.996932672 1
222299.648252619 1
268269.579527973 1
323745.754281764 1
390693.993705461 1
471486.636345739 1
568986.602901829 1
686648.8450043 1
828642.772854684 1
1000000 1
};
\addplot [very thick, darkorange25512714]
table {%
100 0.659848484848485
120.679264063933 0.674074074074074
145.634847750124 0.672254641909814
175.751062485479 0.814195488721804
212.095088792019 0.830161878216123
255.954792269954 0.878815668806498
308.884359647748 0.892278630460449
372.759372031494 0.93706288319931
449.843266896944 0.954127535419295
542.867543932386 0.959976202449165
655.128556859551 0.966061562746646
790.60432109077 0.963254593175853
954.095476349994 0.963043478260869
1151.39539932645 0.963063063063063
1389.49549437314 0.964591874767052
1676.83293681101 0.965432098765432
2023.58964772516 0.965217391304348
2442.05309454865 0.964376590330789
2947.05170255181 0.96414762741652
3556.48030622313 0.964322120285423
4291.93426012878 0.964651948365303
5179.47467923121 0.964614154338265
6250.55192527397 0.965060440470277
7543.12006335462 0.964797913950456
9102.98177991522 0.964387302309705
10985.4114198756 0.964558393816571
13257.1136559011 0.96441823223841
15998.5871960606 0.964631265572922
19306.9772888325 0.964631432172258
23299.5181051537 0.964603932896345
28117.6869797423 0.964649998158854
33932.2177189533 0.964856619890177
40949.1506238042 0.964744845719198
49417.1336132384 0.96468788954883
59636.2331659464 0.9646887260859
71968.5673001151 0.964720098395322
86851.1373751352 0.964703778757897
104811.313415469 0.964747831927461
126485.52168553 0.964714259989851
152641.796717523 0.964770272607352
184206.996932672 0.964652056857973
222299.648252619 0.964734208399988
268269.579527973 0.96469603004002
323745.754281764 0.964691849160868
390693.993705461 0.964704728869409
471486.636345739 0.964737449401824
568986.602901829 0.964695992510692
686648.8450043 0.964719859901681
828642.772854684 0.964722256406284
1000000 0.964758837983205
};
\addplot [very thick, forestgreen4416044]
table {%
100 0.56141592920354
120.679264063933 0.666666666666667
145.634847750124 0.686275862068966
175.751062485479 0.814195488721804
212.095088792019 0.841657417684055
255.954792269954 0.891187244128421
308.884359647748 0.925279106858054
372.759372031494 0.948856194218527
449.843266896944 0.963188588642713
542.867543932386 0.976622797389447
655.128556859551 0.982128982128982
790.60432109077 0.982002805689678
954.095476349994 0.980808426956226
1151.39539932645 0.98141592920354
1389.49549437314 0.98316251830161
1676.83293681101 0.982392228293868
2023.58964772516 0.984437751004016
2442.05309454865 0.983558792924037
2947.05170255181 0.984318455971049
3556.48030622313 0.983709631323235
4291.93426012878 0.983657034580767
5179.47467923121 0.983912105159898
6250.55192527397 0.98407022106632
7543.12006335462 0.983765577635567
9102.98177991522 0.983868266815518
10985.4114198756 0.983764281419122
13257.1136559011 0.984099007624813
15998.5871960606 0.984060455959865
19306.9772888325 0.984081879653748
23299.5181051537 0.984018314618991
28117.6869797423 0.984230776178176
33932.2177189533 0.98419350975931
40949.1506238042 0.983996625645097
49417.1336132384 0.984223887193085
59636.2331659464 0.98418430353873
71968.5673001151 0.984170001340927
86851.1373751352 0.984133482270789
104811.313415469 0.984108983585583
126485.52168553 0.984056224899599
152641.796717523 0.984100045257301
184206.996932672 0.984251447477254
222299.648252619 0.984140548420338
268269.579527973 0.984077753730623
323745.754281764 0.984150941028071
390693.993705461 0.984170772610861
471486.636345739 0.984098460265543
568986.602901829 0.984168528623854
686648.8450043 0.984157605821231
828642.772854684 0.984160025302769
1000000 0.984135240043763
};
\legend{$\mathrm{GT}\leftrightarrow\mathrm{GT}$,$\mathrm{GT}\leftrightarrow\mathrm{1}$,$\mathrm{GT}\leftrightarrow\mathrm{2}$}
\end{axis}

\end{tikzpicture}
    \end{subfigure}
    \caption{Visualization of the effect of varying number of samples on the $F_{1\mathrm{cm}}$ metric. Note that compared to CD (see Fig.\ \ref{fig:cd}) $F_{1\mathrm{cm}}$ converges significantly faster.}
    \label{fig:fscore}
\end{figure}
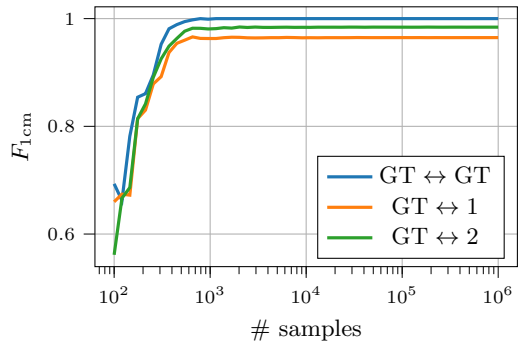

In Fig.\ \ref{fig:fscore} we show $F_{1\mathrm{cm}}$ for varying numbers of samples for the meshes in Fig.\ \ref{fig:example_meshes}. $F_{1\mathrm{cm}}$ converges significantly faster than CD and can easily be interpreted as the percentage of correct (i.e., error below $\Delta=\SI{1}{cm}$) surfaces or points \cite{tatarchenko2019single}.

All metrics so far (i.e., $d,\delta,\mathrm{IoU},\mathrm{CD},\text{ and }F_\Delta$) assess the quality of a single estimate. Next, we discuss average precision and precision, which attempt to summarize a method's performance on a dataset. In principle, one could also compute different averages of the aforementioned metrics, but those are typically affected by outliers, and hard to interpret in comparison to thresholded evaluation metrics that classify estimates as true or false.

\subsubsection{Average precision}\label{section:ap}

\emph{Average precision} (AP) summarizes precision-recall curves in a single value \cite{salton1983introduction} and has been the standard evaluation metric for object detection on the PASCAL VOC \cite{everingham2010pascal} and COCO datasets \cite{lin2014microsoft}. In general, average precision is calculated based on the interpolated precision-recall curve, which is constructed by varying a confidence threshold. 

Wang et al.\ \cite{wang2019normalized} proposed to use AP with different thresholds on $\mathrm{IoU}^+$, $d$ and $\delta$ to evaluate their pose estimation (all specified thresholds have to hold for a prediction to count as true positive). Their method includes a Mask R-CNN architecture to detect objects and therefore had a confidence threshold to work with. However, \emph{none} of the following pose and shape estimation methods include such a confidence threshold. Instead they all assume $\mathbf{M}$ to be given as stated in Problem \ref{problem}. 

To still follow the same evaluation protocol as \cite{wang2019normalized}, all other methods rely on the same, suboptimal Mask R-CNN predictions that \cite{wang2019normalized} provided. This protocol effectively limits the achievable AP due to wrong classifications, missing detections and poor masks. Furthermore, AP is inherently difficult to interpret compared to simpler metrics.

Therefore, we believe that AP is unnecessary to compare pose and shape estimation methods. Rather, simpler metrics such as precision (see below) should be used, by assuming $\mathbf{M}$ and $c$ to be given.

\subsubsection{Precision}\label{section:precision} We propose to use precision, contrary to average precision to assess categorical pose and shape estimation:
\begin{definition}
    Given inputs $(\mathbf{I}^i,\mathbf{D}^i,\mathbf{P}^i,\mathbf{M}^i,c^i)$, ground truths $(\mathcal{O}^i,\tensor*[^{\mathrm{w}}]{\mathbf{T}}{^i_{\mathrm{o}}})$ and associated predictions $(\widetilde{\mathcal{O}}^i,\tensor*[^{\mathrm{w}}]{\widetilde{\mathbf{T}}}{^i_{\mathrm{o}}})$ with $i=1,...,N$, let precision be defined as
    \begin{equation}
        P=\frac{\sum_{i=1}^N\left[ \mathrm{c}(\mathcal{O}^i,\tensor*[^{\mathrm{w}}]{\mathbf{T}}{^i_{\mathrm{o}}}, \widetilde{\mathcal{O}}^i,\tensor*[^{\mathrm{w}}]{\widetilde{\mathbf{T}}}{^i_{\mathrm{o}}}) \right]}{N},
    \end{equation}
    where $\left[\cdot\right]$ denotes the Iverson bracket and $\mathrm{c}$ determines whether a prediction is correct or not based on a single or multiple thresholds on $\mathrm{IoU}$, $\delta$, $d$ or $F_\Delta$.
\end{definition}
That is, precision measures the percentage of correct estimates based on thresholds on translation error, rotation error, IoU and F-score. Note that we can use this simpler metric instead of average precision, because we decouple pose estimation from detection and classification.


\subsubsection{Summary}\label{section:metrics_summary}

We propose to evaluate categorical pose and shape estimation methods by calculating precision at varying thresholds of $d$, $\delta$ and $F_\mathrm{\Delta}$. Furthermore, this evaluation procedure can be adjusted for categorical pose estimation by using $d$ and $\delta$ only, and for categorical pose and size estimation by using $\mathrm{IoU}$ instead of $F_\mathrm{\Delta}$. The thresholds for these metrics must be adjusted based on the application requirements, sensor accuracy, and annotation quality. Furthermore, when combining multiple thresholds, care should be taken that they are roughly equally strict.

Note that when evaluating $\delta$ and $\mathrm{IoU}$, extra care has to be taken of symmetric object categories. We follow \cite{wang2019normalized} and ignore rotations around the up-axis for bottle, bowl and can categories. Further issues with ambiguities are discussed in Section \ref{sec:limitations}.

\subsection{Datasets}\label{section:datasets}

So far, most methods evaluate on the synthetic CAMERA25 dataset and on the smaller real-world dataset REAL275 \cite{wang2019normalized}. Since we are most interested in real-world performance we only include the REAL275 dataset in our evaluation protocol. Next, we will give an overview over the REAL275 dataset, and our new annotations for the Redwood dataset \cite{choi2016large}.

\subsubsection{REAL275}\label{section:real275}

The REAL dataset was proposed by Wang et al.\ \cite{wang2019normalized} and consists of 4300 training images (7 video sequences) and 2750 test images (6 video sequences). 
The dataset contains 6 categories (bottle, bowl, camera, can, laptop, and mug) and contains 4 to 7 objects per scene. Meshes for each object are provided, obtained using an RGB-D reconstruction algorithm. Since we are primarily interested in evaluation, we will focus on the evaluation split, referred to as REAL275, from now.

\begin{figure}
    \centering
    \includegraphics[width=0.3\textwidth]{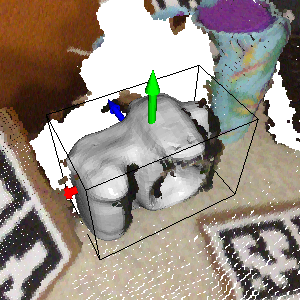}
    \includegraphics[width=0.3\textwidth]{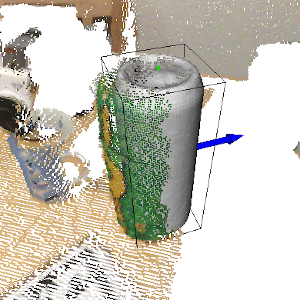}
    \includegraphics[width=0.3\textwidth]{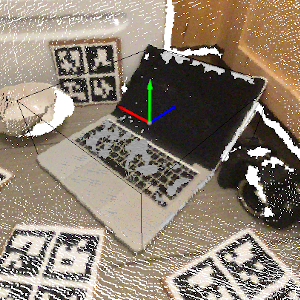}
    \includegraphics[width=0.3\textwidth]{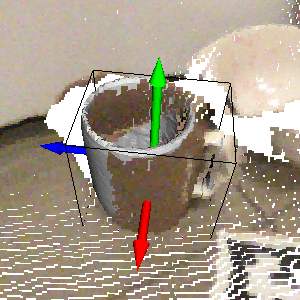}
    \includegraphics[width=0.3\textwidth]{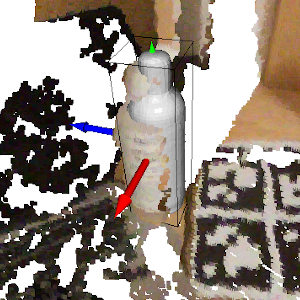}
    \includegraphics[width=0.3\textwidth]{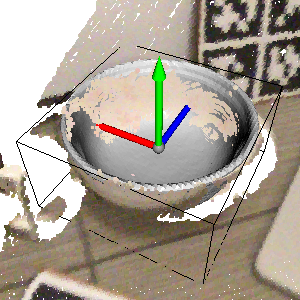}
    \caption{Examples of REAL275 samples for the 6 object categories. Note that all objects are positioned upright on a table.}
    \label{fig:real275_examples}
\end{figure}

Fig.\ \ref{fig:real275_examples} shows point sets with their corresponding ground truth annotations. Note that all objects are upright on a planar surfaces. Similar constrained orientations can be found in the training splits of the CAMERA and REAL datasets. Fig.\ \ref{fig:orientation_real275} visualizes the distribution of orientations contained in the REAL275 dataset. Note that such constraints, present in training and test data, can significantly simplify the learning problem as pose and shape ambiguities disappear (e.g., upright or upside down can).


\subsubsection{Redwood}\label{section:redwood}

To evaluate methods on less constrained orientations, we contribute annotations for a set of images in the Redwood dataset \cite{choi2016large}. The Redwood dataset contains sequences of handheld objects being freely rotated in front of the camera. No ground truth reconstruction is provided for those sequences. 

We annotated pose and shape for 3 categories (bottle, bowl, mug) for 5 sequences each. These annotations were created by manually creating OBBs in multiple frames and exploiting potential symmetries of the object. Alignment of OBBs with previous annotations was sped up and refined by using the iterative closest point (ICP) algorithm. For each of the annotated sequences we took a subset of 5 frames covering various orientations. We will refer to this set of annotations as REDWOOD75.

\begin{figure}[tb]
    \centering
    \includegraphics[scale=0.7]{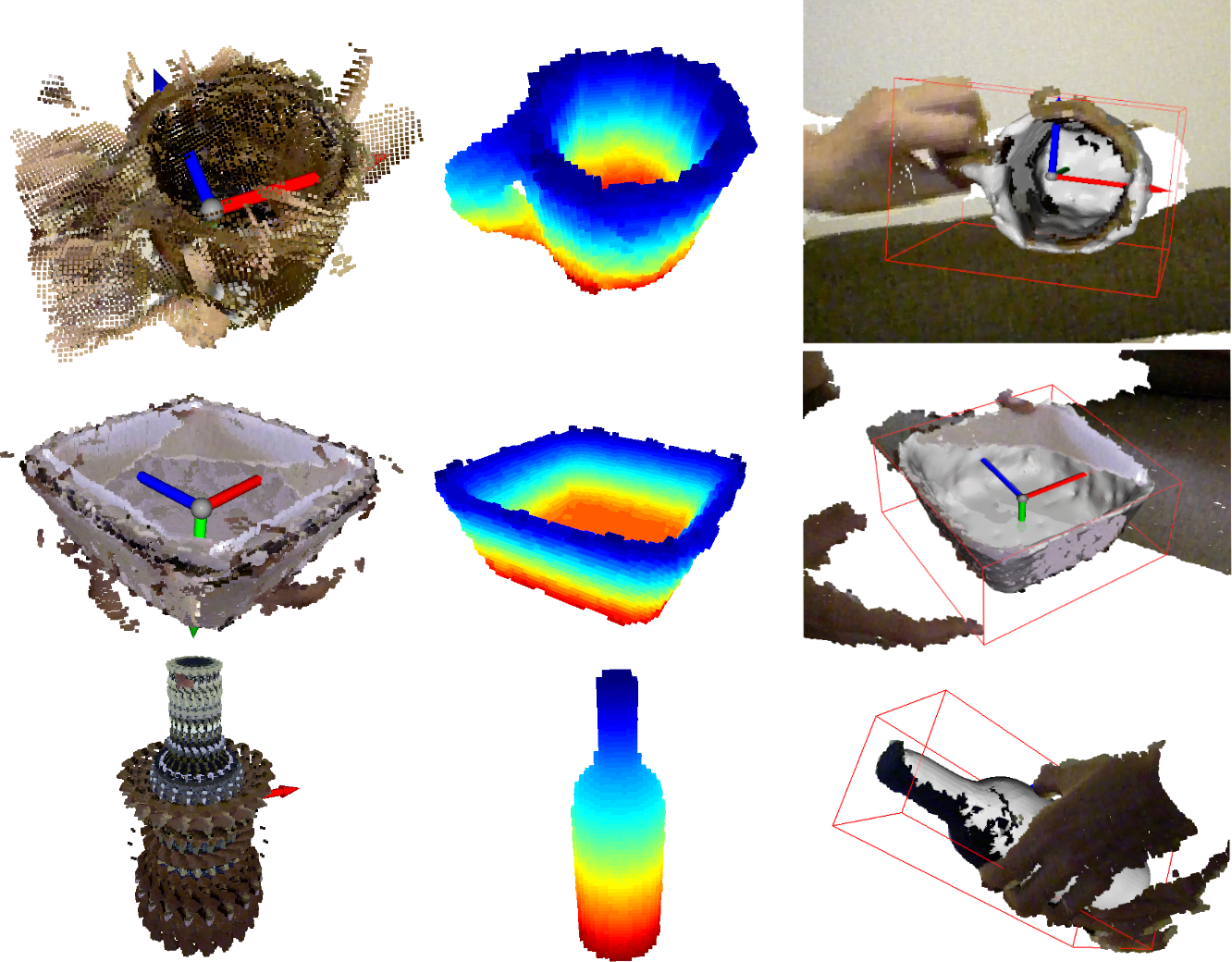}
    \caption{Manual annotations for Redwood dataset. The left column shows the cropped, accumulated point sets (including symmetries) extracted from annotated bounding boxes. The middle columns shows the voxel grid after carving. The right column shows the extracted mesh, overlaid with the point set.}\label{fig:redwood_annotations}
\end{figure}

To reconstruct the shape from the occluded and noisy depth data, we start from a dense voxel grid inside the bounding box and apply voxel carving using the annotated frames to remove hands and other temporary occlusions. The remaining voxel grid only contains the voxels that are not observed as free in any of the annotated frames. From this voxel grid, we extract a mesh and apply Laplacian smoothing. Fig.\ \ref{fig:redwood_annotations} visualizes the annotation process.

We note that this method only approximates the real shape and is sensitive to misaligned bounding boxes and missing depth data. Especially thin surfaces and details such as mug handles are difficult to extract accurately due to sensor noise. Also, alignment errors can easily accumulate, resulting in too big or too small objects. However, the annotations are accurate enough to evaluate current methods on unconstrained orientations.

\begin{figure}[htb]
    \centering
    \includegraphics[width=0.25\textwidth]{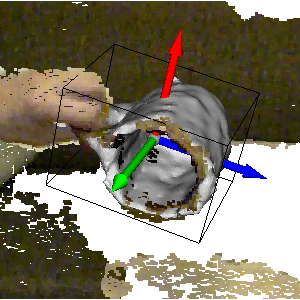}
    \includegraphics[width=0.25\textwidth]{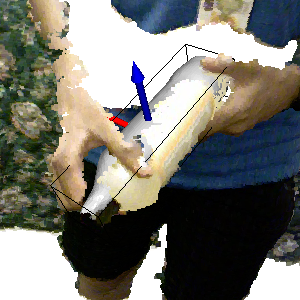}
    \includegraphics[width=0.25\textwidth]{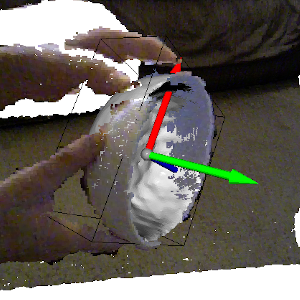}
    \includegraphics[width=0.25\textwidth]{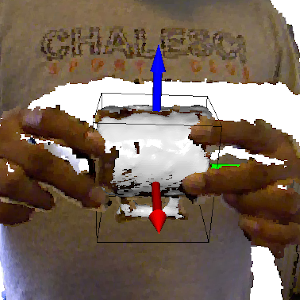}
    \includegraphics[width=0.25\textwidth]{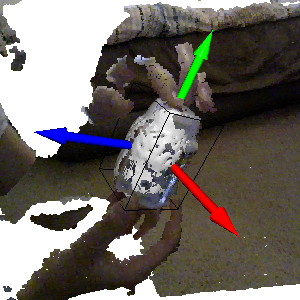}
    \includegraphics[width=0.25\textwidth]{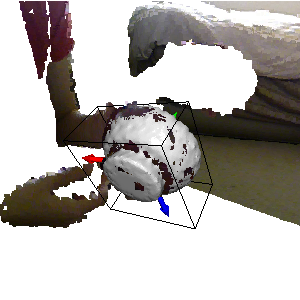}
    \caption{Examples of REDWOOD75 samples.}
    \label{fig:redwood_examples}
\end{figure}

\begin{figure}[htb]
    \centering
    \begin{subfigure}{0.49\textwidth}
        \includegraphics[width=0.49\textwidth]{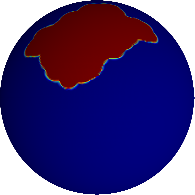}
        \includegraphics[width=0.49\textwidth]{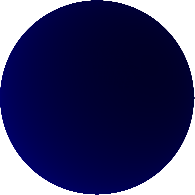}
        \caption{REAL275}\label{fig:orientation_real275}
    \end{subfigure}
    \begin{subfigure}{0.49\textwidth}
        \includegraphics[width=0.49\textwidth]{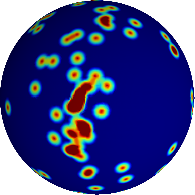}
        \includegraphics[width=0.49\textwidth]{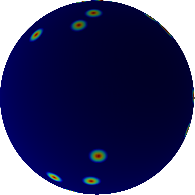}
        \caption{REDWOOD75}\label{fig:orientation_redwood75}
    \end{subfigure}
    \caption{Distribution of the up-axis in REAL275 and REDWOOD75 datasets. REDWOOD75 covers a larger variety of orientations.}
    \label{fig:orientation_dist}
\end{figure}

To produce the final ground truth, we compute tight bounding boxes based on the extracted meshes. Fig.\ \ref{fig:redwood_examples} shows examples of the final annotations. In Fig.\ \ref{fig:orientation_dist} we compare the orientation distribution of REDWOOD75 and REAL275. Note that compared to REAL275, the orientations are significantly less constrained in REDWOOD75.

\section{Experiments}\label{sec:experiments}

We follow our proposed evaluation protocol and compare three methods for shape and pose estimation: CASS \cite{chen2020learning}, SPD \cite{tian2020shape} and ASM-Net \cite{akizuki2021}. CASS and SPD are well-established baselines for this task and were both trained on the CAMERA and REAL datasets. In contrast to the other methods, ASM-Net \cite{akizuki2021} was trained on synthetic ShapeNet \cite{chang2015} renderings only. All methods estimate 6D pose and reconstruct a point set of varying density.

For all methods we closely followed the published inference code and verified that our method interface produced similar results as their evaluation code. We found that only SPD's published model achieved the same qualitative results as shown in their publication. CASS' reconstructed point sets were significantly worse except for the laptop category and ASM-Net often predicted negative scales, which causes some reconstructions to be upside down, while the object frame $\mathrm{o}$ is predicted in the correct orientation.

We have implemented the metrics, and interfaces to the datasets and methods described in the previous sections using Open3D \cite{zhou2018open3d} and PyTorch \cite{paszke2019pytorch}. We open source our code as a benchmarking toolbox, with the goal of simplifying fair comparison with state-of-the-art methods. We plan to extend the toolbox as new methods are released. 

\paragraph{Qualitative Results}\label{sec:qualitative}

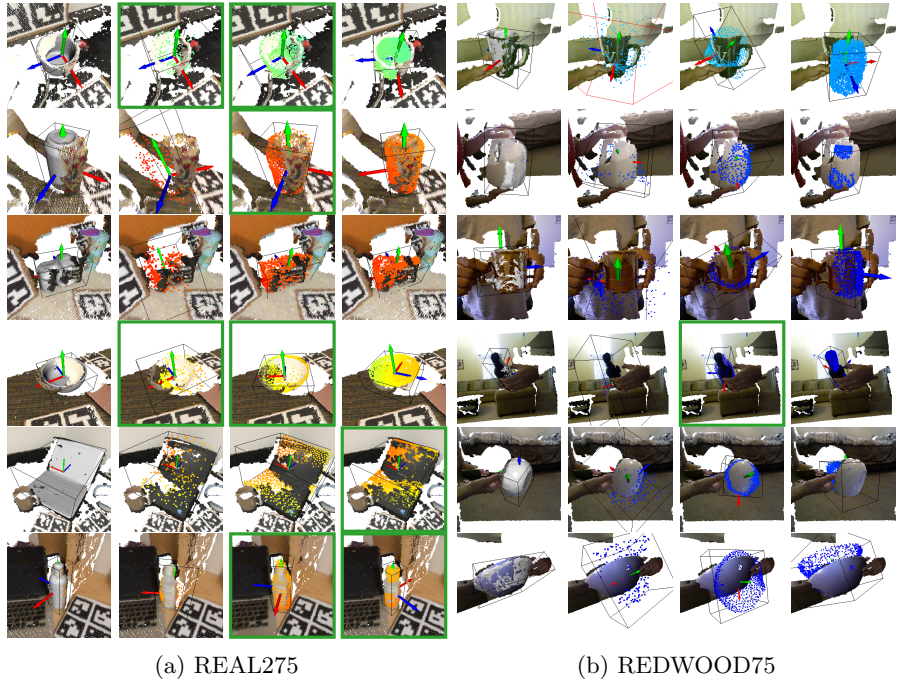
\begin{figure}[htb]
    \centering
    \begin{subfigure}{0.49\textwidth}
        \centering
        \includegraphics[trim={4cm 0 4cm 0},clip,width=0.23\textwidth,height=0.23\textwidth]{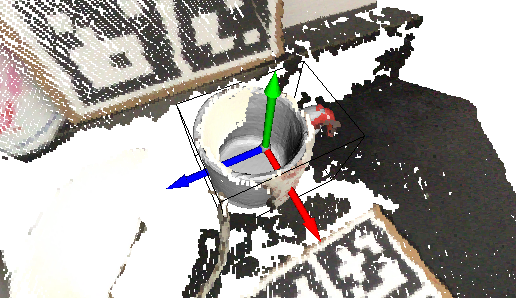}
        \includegraphics[trim={4cm 0 4cm 0},clip,width=0.23\textwidth,height=0.23\textwidth]{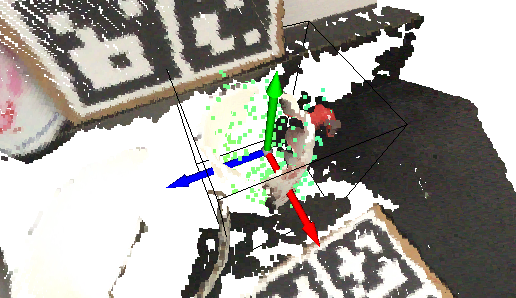}%
        \begin{tikzpicture}[overlay]%
            \draw[forestgreen4416044, very thick] (0,0) rectangle ++(-0.23\textwidth,0.23\textwidth);%
        \end{tikzpicture}
        \includegraphics[trim={4cm 0 4cm 0},clip,width=0.23\textwidth,height=0.23\textwidth]{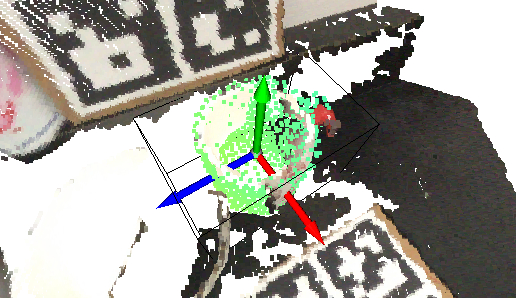}%
        \begin{tikzpicture}[overlay]%
            \draw[forestgreen4416044, very thick] (0,0) rectangle ++(-0.23\textwidth,0.23\textwidth);%
        \end{tikzpicture}
        \includegraphics[trim={4cm 0 4cm 0},clip,width=0.23\textwidth,height=0.23\textwidth]{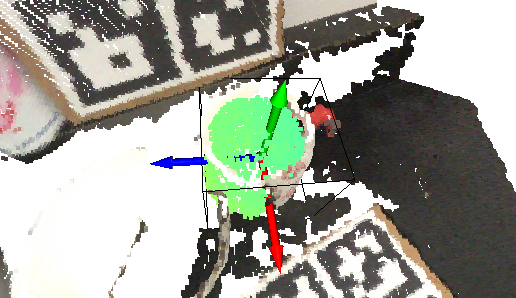}
        
        \includegraphics[trim={4cm 0 4cm 0},clip,width=0.23\textwidth,height=0.23\textwidth]{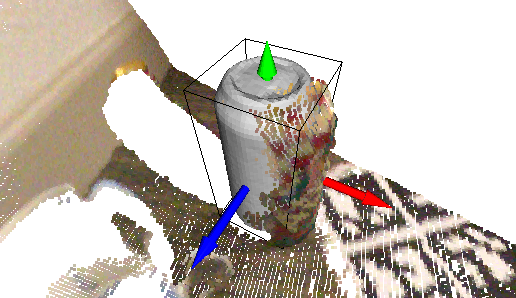}
        \includegraphics[trim={4cm 0 4cm 0},clip,width=0.23\textwidth,height=0.23\textwidth]{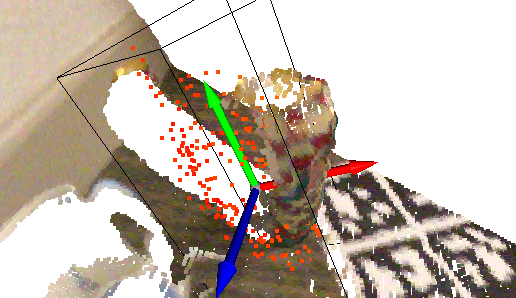}
        \includegraphics[trim={4cm 0 4cm 0},clip,width=0.23\textwidth,height=0.23\textwidth]{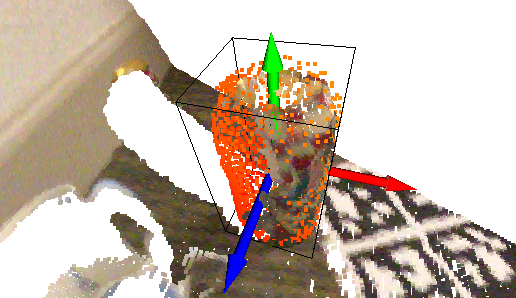}%
        \begin{tikzpicture}[overlay]%
            \draw[forestgreen4416044, very thick] (0,0) rectangle ++(-0.23\textwidth,0.23\textwidth);%
        \end{tikzpicture}
        \includegraphics[trim={4cm 0 4cm 0},clip,width=0.23\textwidth,height=0.23\textwidth]{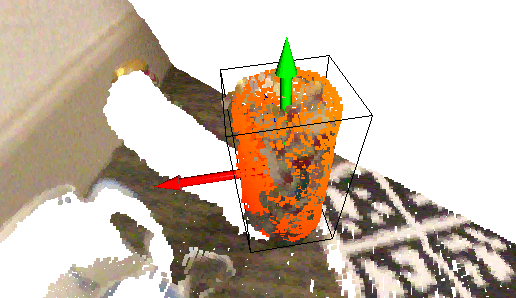}
        
        \includegraphics[trim={4cm 0 4cm 0},clip,width=0.23\textwidth,height=0.23\textwidth]{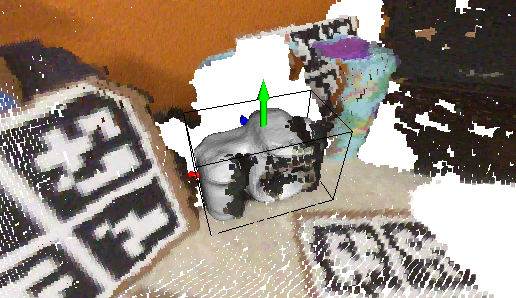}
        \includegraphics[trim={4cm 0 4cm 0},clip,width=0.23\textwidth,height=0.23\textwidth]{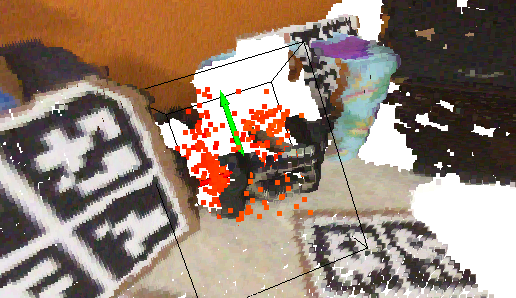}
        \includegraphics[trim={4cm 0 4cm 0},clip,width=0.23\textwidth,height=0.23\textwidth]{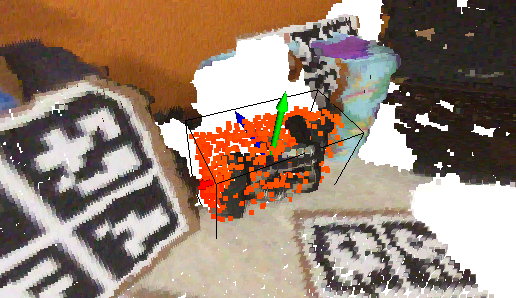}
        \includegraphics[trim={4cm 0 4cm 0},clip,width=0.23\textwidth,height=0.23\textwidth]{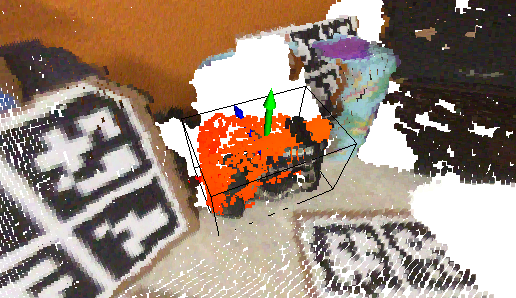}
        
        \includegraphics[trim={4cm 0 4cm 0},clip,width=0.23\textwidth,height=0.23\textwidth]{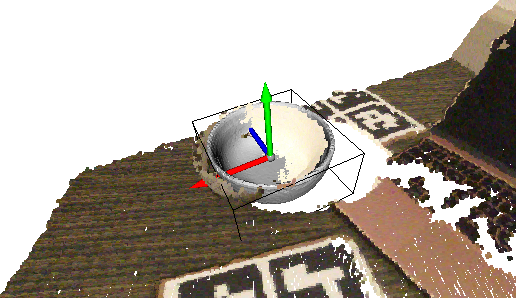}
        \includegraphics[trim={4cm 0 4cm 0},clip,width=0.23\textwidth,height=0.23\textwidth]{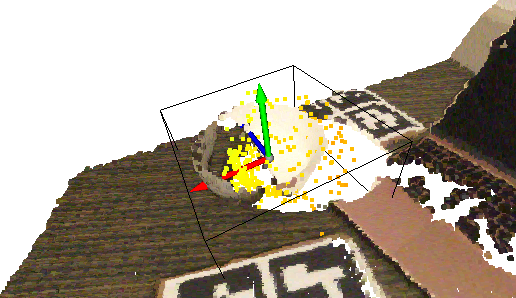}%
        \begin{tikzpicture}[overlay]%
            \draw[forestgreen4416044, very thick] (0,0) rectangle ++(-0.23\textwidth,0.23\textwidth);%
        \end{tikzpicture}
        \includegraphics[trim={4cm 0 4cm 0},clip,width=0.23\textwidth,height=0.23\textwidth]{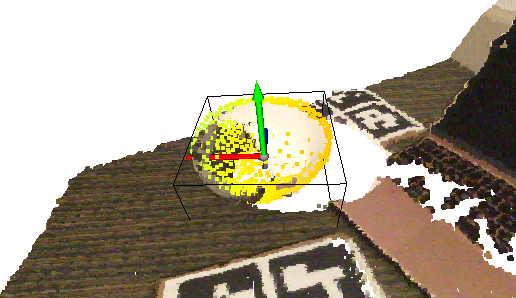}%
        \begin{tikzpicture}[overlay]%
            \draw[forestgreen4416044, very thick] (0,0) rectangle ++(-0.23\textwidth,0.23\textwidth);%
        \end{tikzpicture}
        \includegraphics[trim={4cm 0 4cm 0},clip,width=0.23\textwidth,height=0.23\textwidth]{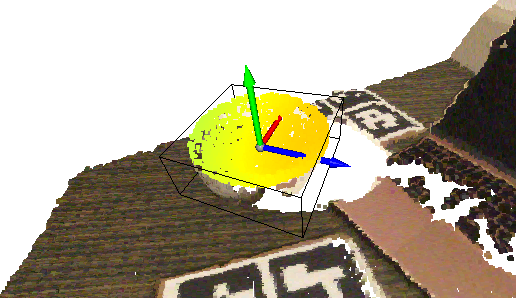}
        
        \includegraphics[trim={4cm 0 4cm 0},clip,width=0.23\textwidth,height=0.23\textwidth]{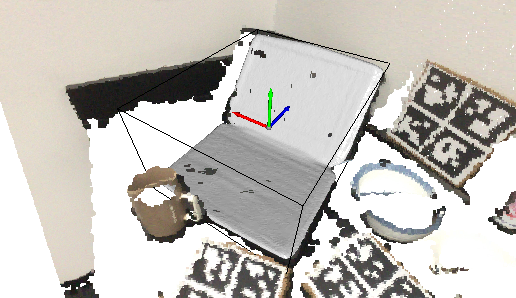}
        \includegraphics[trim={4cm 0 4cm 0},clip,width=0.23\textwidth,height=0.23\textwidth]{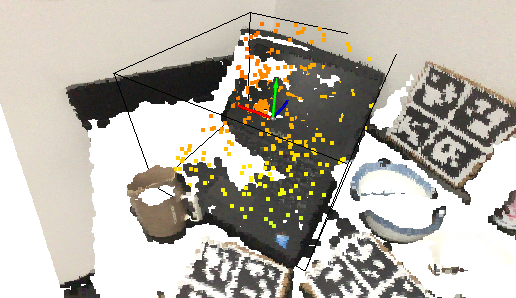}
        \includegraphics[trim={4cm 0 4cm 0},clip,width=0.23\textwidth,height=0.23\textwidth]{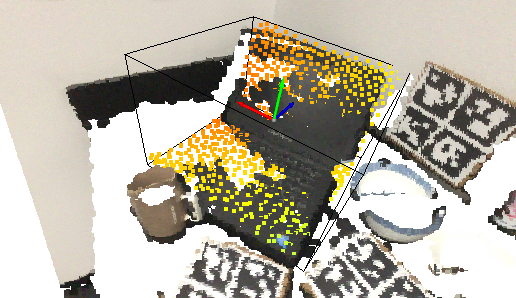}
        \includegraphics[trim={4cm 0 4cm 0},clip,width=0.23\textwidth,height=0.23\textwidth]{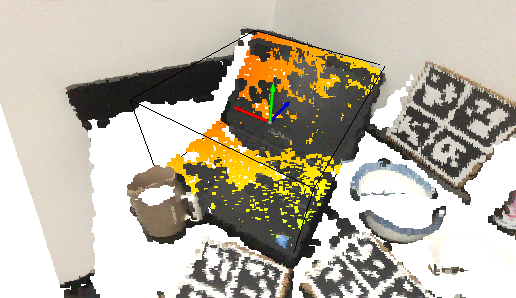}%
        \begin{tikzpicture}[overlay]%
            \draw[forestgreen4416044, very thick] (0,0) rectangle ++(-0.23\textwidth,0.23\textwidth);%
        \end{tikzpicture}
        
        \includegraphics[trim={4cm 0 4cm 0},clip,width=0.23\textwidth,height=0.23\textwidth]{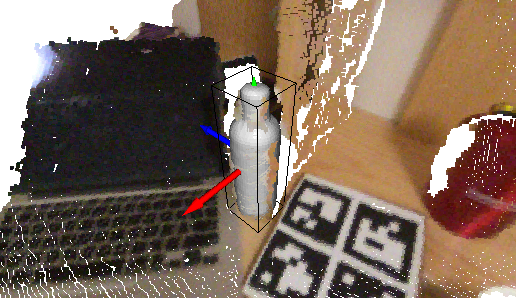}
        \includegraphics[trim={4cm 0 4cm 0},clip,width=0.23\textwidth,height=0.23\textwidth]{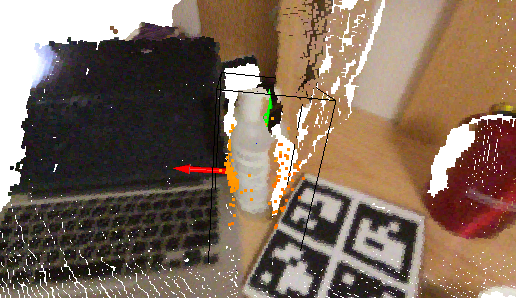}
        \includegraphics[trim={4cm 0 4cm 0},clip,width=0.23\textwidth,height=0.23\textwidth]{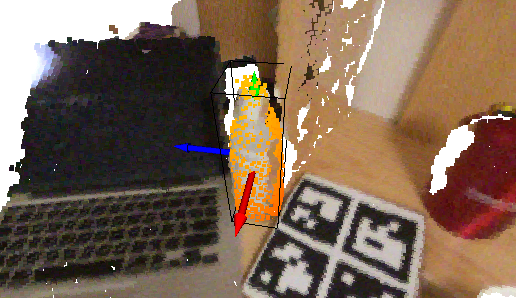}%
        \begin{tikzpicture}[overlay]%
            \draw[forestgreen4416044, very thick] (0,0) rectangle ++(-0.23\textwidth,0.23\textwidth);%
        \end{tikzpicture}
        \includegraphics[trim={4cm 0 4cm 0},clip,width=0.23\textwidth,height=0.23\textwidth]{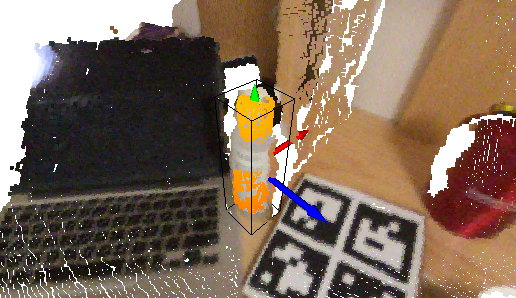}%
        \begin{tikzpicture}[overlay]%
            \draw[forestgreen4416044, very thick] (0,0) rectangle ++(-0.23\textwidth,0.23\textwidth);%
        \end{tikzpicture}
        \caption{REAL275}
    \end{subfigure}%
    \begin{subfigure}{0.49\textwidth}
        \centering
        \includegraphics[trim={4cm 0 4cm 0},clip,width=0.23\textwidth,height=0.23\textwidth]{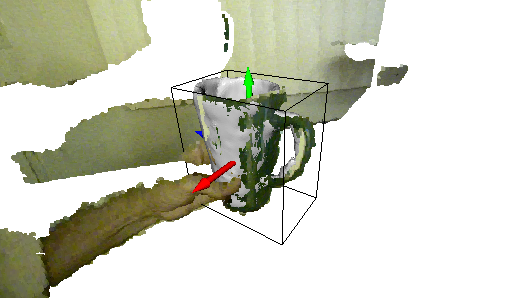}
        \includegraphics[trim={4cm 0 4cm 0},clip,width=0.23\textwidth,height=0.23\textwidth]{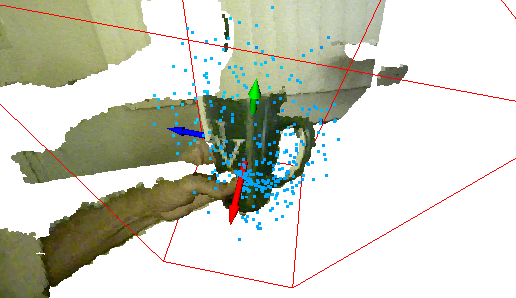}
        \includegraphics[trim={4cm 0 4cm 0},clip,width=0.23\textwidth,height=0.23\textwidth]{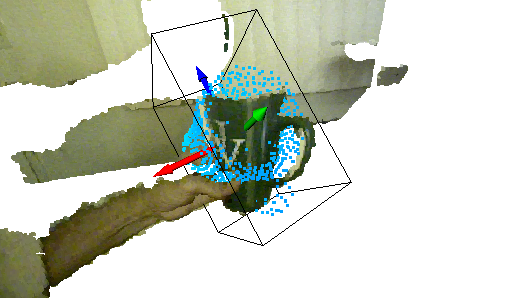}
        \includegraphics[trim={4cm 0 4cm 0},clip,width=0.23\textwidth,height=0.23\textwidth]{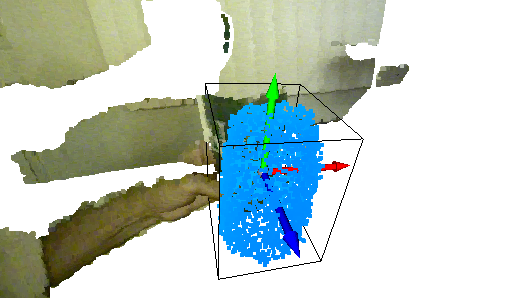}
        
        \includegraphics[trim={4cm 0 4cm 0},clip,width=0.23\textwidth,height=0.23\textwidth]{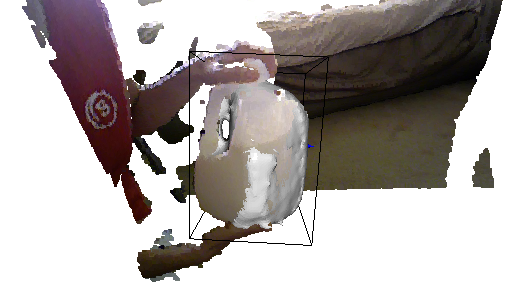}
        \includegraphics[trim={4cm 0 4cm 0},clip,width=0.23\textwidth,height=0.23\textwidth]{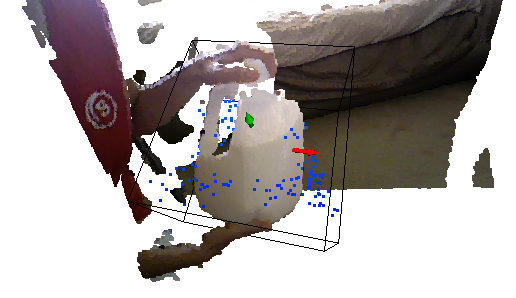}
        \includegraphics[trim={4cm 0 4cm 0},clip,width=0.23\textwidth,height=0.23\textwidth]{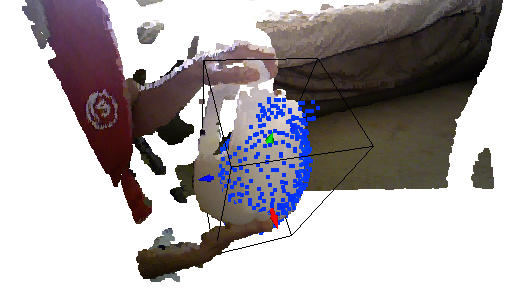}
        \includegraphics[trim={4cm 0 4cm 0},clip,width=0.23\textwidth,height=0.23\textwidth]{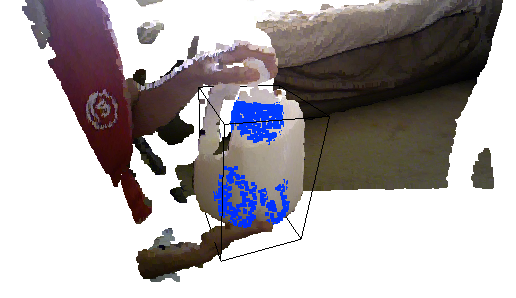}
        
        \includegraphics[trim={4cm 0 4cm 0},clip,width=0.23\textwidth,height=0.23\textwidth]{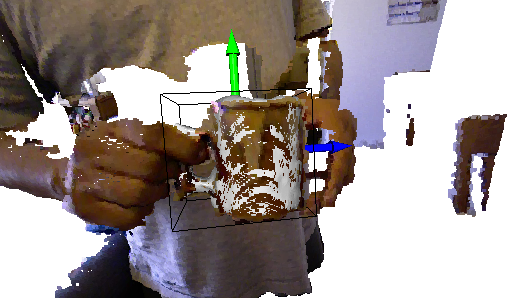}
        \includegraphics[trim={4cm 0 4cm 0},clip,width=0.23\textwidth,height=0.23\textwidth]{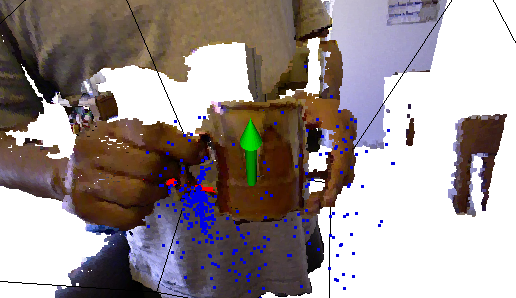}
        \includegraphics[trim={4cm 0 4cm 0},clip,width=0.23\textwidth,height=0.23\textwidth]{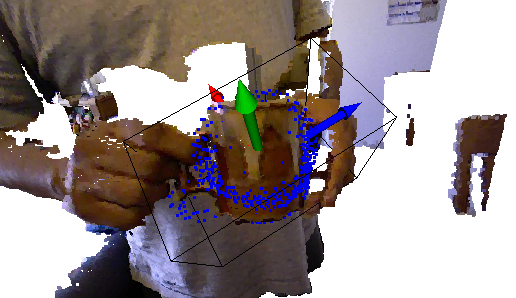}
        \includegraphics[trim={4cm 0 4cm 0},clip,width=0.23\textwidth,height=0.23\textwidth]{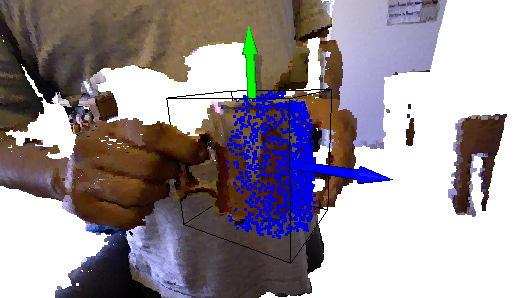}
        
        \includegraphics[trim={4cm 0 4cm 0},clip,width=0.23\textwidth,height=0.23\textwidth]{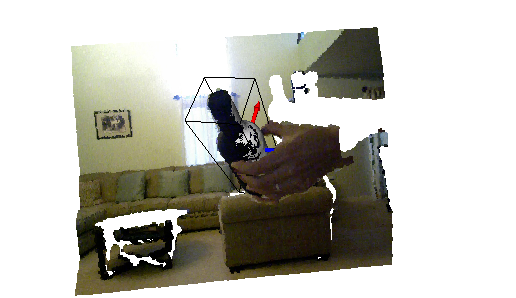}
        \includegraphics[trim={4cm 0 4cm 0},clip,width=0.23\textwidth,height=0.23\textwidth]{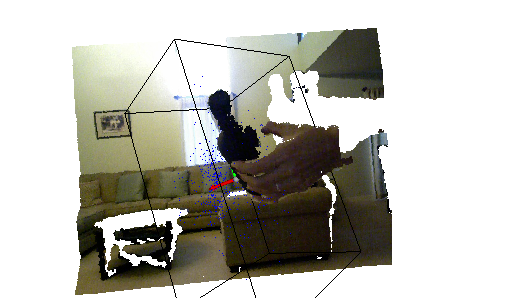}
        \includegraphics[trim={4cm 0 4cm 0},clip,width=0.23\textwidth,height=0.23\textwidth]{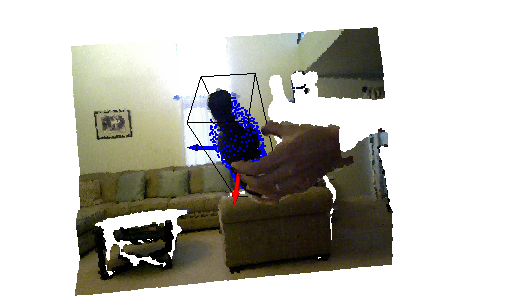}%
        \begin{tikzpicture}[overlay]%
            \draw[forestgreen4416044, very thick] (0,0) rectangle ++(-0.23\textwidth,0.23\textwidth);%
        \end{tikzpicture}
        \includegraphics[trim={4cm 0 4cm 0},clip,width=0.23\textwidth,height=0.23\textwidth]{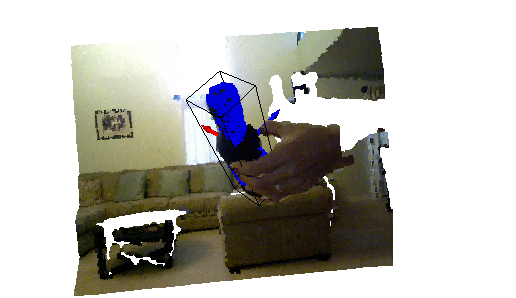}
        
        \includegraphics[trim={4cm 0 4cm 0},clip,width=0.23\textwidth,height=0.23\textwidth]{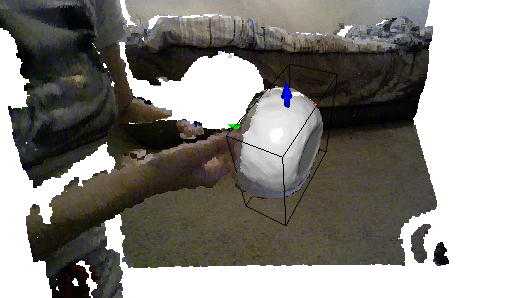}
        \includegraphics[trim={4cm 0 4cm 0},clip,width=0.23\textwidth,height=0.23\textwidth]{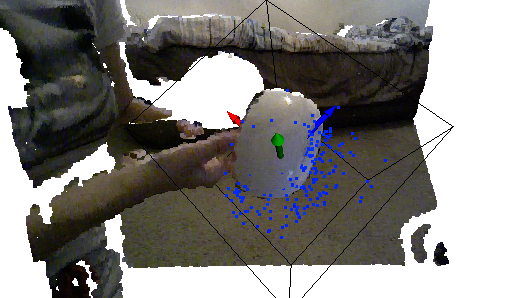}
        \includegraphics[trim={4cm 0 4cm 0},clip,width=0.23\textwidth,height=0.23\textwidth]{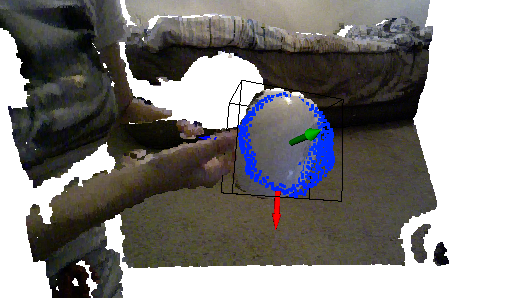}
        \includegraphics[trim={4cm 0 4cm 0},clip,width=0.23\textwidth,height=0.23\textwidth]{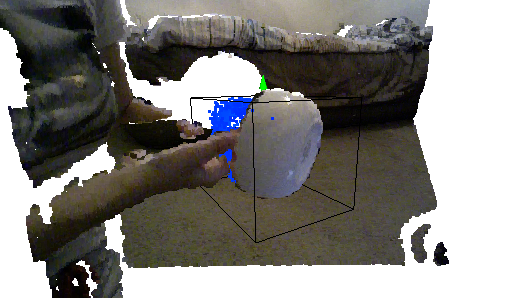}
        
        \includegraphics[trim={4cm 0 4cm 0},clip,width=0.23\textwidth,height=0.23\textwidth]{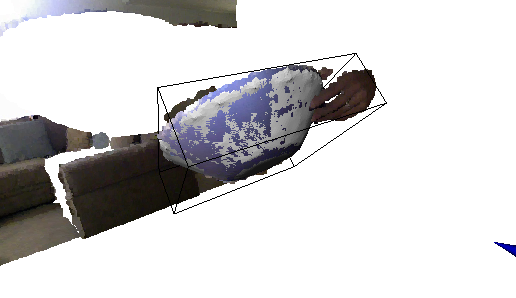}
        \includegraphics[trim={4cm 0 4cm 0},clip,width=0.23\textwidth,height=0.23\textwidth]{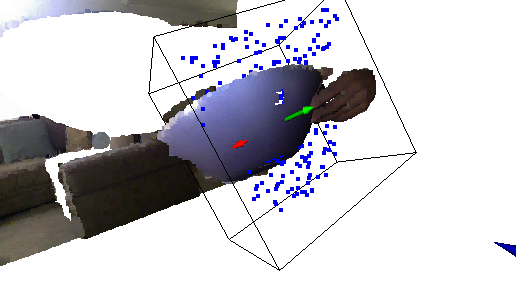}
        \includegraphics[trim={4cm 0 4cm 0},clip,width=0.23\textwidth,height=0.23\textwidth]{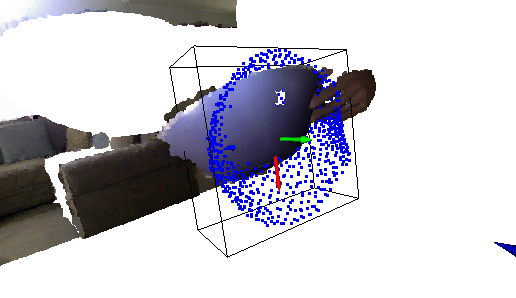}
        \includegraphics[trim={4cm 0 4cm 0},clip,width=0.23\textwidth,height=0.23\textwidth]{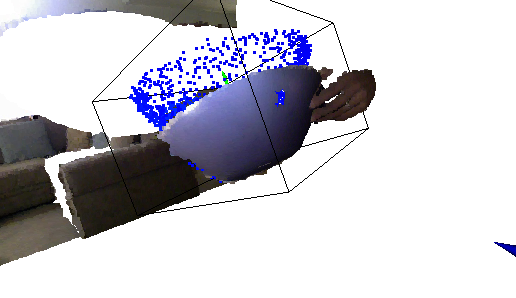}
        \caption{REDWOOD75}
    \end{subfigure}
    \caption{Randomly selected results on REAL275 and REDWOOD75 datasets. Results that are considered correct under $\delta=10^{\circ},d=\SI{2}{cm},F_{1\mathrm{cm}}=0.6$ thresholds are highlighted.}
    \label{fig:qualitative}
\end{figure}

Fig.\ \ref{fig:qualitative} shows randomly picked results on the REAL275 and REDWOOD75 datasets. On REAL275, all methods perform pose estimation with a similar quality as shown in the respective publications. On REDWOOD75, on the other hand, only ASM-Net shows limited generalization capability. CASS and SPD predict upright objects consistent with the orientation distribution of REAL275 independent of the input.

The shape reconstructions of SPD are qualitatively the best. ASM-Net's reconstructions are often flipped, but surfaces typically align well. As noted before, CASS completely fails to reconstruct any object except laptops. For both ASM-Net and CASS it is unclear, whether this performance difference stems from errors in the code or if the published model weights are suboptimal.

\paragraph{Quantitative Results}\label{sec:quantitative}

\begin{table}[htb]
    \centering
    \caption{Precision at varying position, orientation and F-score thresholds.}\label{tab:precision}
    \scriptsize
    \begin{tabular}{@{}lR{1.3cm}R{1.3cm}R{1.3cm}cR{1.3cm}R{1.3cm}R{1.3cm}@{}}
        \toprule
         & \multicolumn{3}{c}{REAL275} & \phantom{a} & \multicolumn{3}{c}{REDWOOD75} \\
         \cmidrule{2-4} \cmidrule{6-8}
         & CASS & SPD & ASM-Net & & CASS & SPD & ASM-Net \\
        \midrule
        10\degree,\SI{2}{cm} & 0.331 & \textbf{0.535} & 0.331 & & 0.013 & 0.2 & \textbf{0.307} \\
        5\degree,\SI{1}{cm} & 0.073 & \textbf{0.205} & 0.069 & & 0.000 & 0.013 & \textbf{0.080} \\
        10\degree,\SI{2}{cm},0.6 & 0.031 & \textbf{0.471} & 0.215 & & 0.000 & \textbf{0.173} & \textbf{0.173} \\
        5\degree,\SI{1}{cm},0.8 & 0.000 & \textbf{0.170} & 0.050 & & 0.000 & 0.013 & \textbf{0.053} \\
        \bottomrule
    \end{tabular}
\end{table}

We now present results using the metrics introduced in Section \ref{section:metrics}. In Table \ref{tab:precision} we report precision with thresholds of varying strictness. To assess pose estimation independent of shape we use $5\degree, \SI{1}{cm}$ and $10\degree, \SI{2}{cm}$. To further include shape reconstruction, we use $5\degree, \SI{1}{cm}, 0.8$ and $10\degree, \SI{2}{cm}, 0.6$ for $\delta, d,$ and $F_{1\mathrm{cm}}$, respectively. We picked these tuples of thresholds such that all thresholds in a tuple are roughly equally strict. In the past some methods used pairs such as $10\degree, 5\mathrm{cm}$, or $10\degree, 10\mathrm{cm}$, where only the $10\degree$ threshold practically mattered.

The results from Table \ref{tab:precision} confirm the qualitative observations from before. CASS in particular performs poorly on the shape reconstruction metrics. Note that for both datasets there is still a lot of room for improvement. Typically, significantly fewer than 50\% of the estimates are of sufficient quality to be considered correct in pose and shape. This shows that categorical pose and shape estimation is still an open problem, especially for unconstrained orientations. The performance gap between the two datasets is especially noticeable for SPD, which performs better than ASM-Net on REAL275, but worse on REDWOOD75.

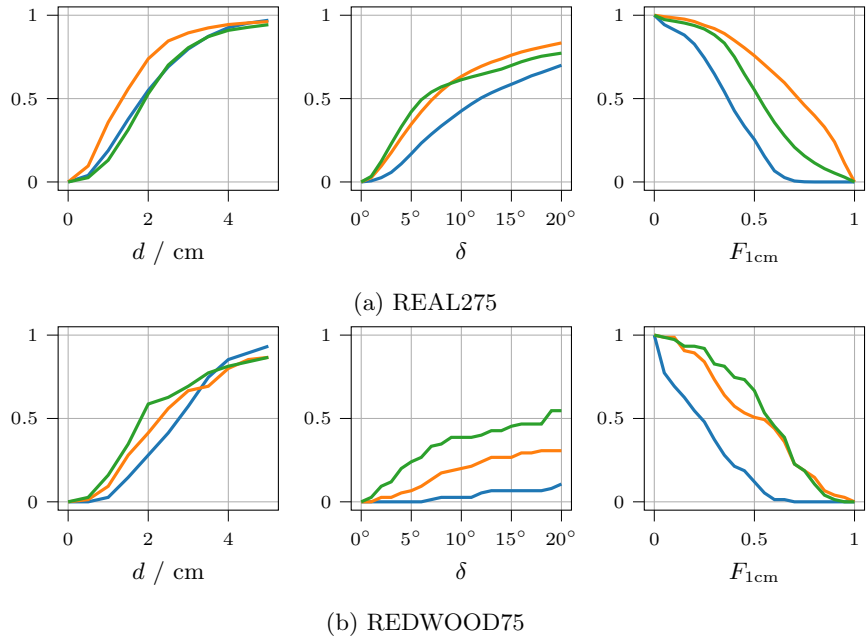
\begin{figure}[htb]
    \centering
    \begin{subfigure}{\textwidth}
        \centering
\begin{tikzpicture}
\definecolor{crimson2143940}{RGB}{214,39,40}
\definecolor{darkgray176}{RGB}{176,176,176}
\definecolor{darkorange25512714}{RGB}{255,127,14}
\definecolor{forestgreen4416044}{RGB}{44,160,44}
\definecolor{lightgray204}{RGB}{204,204,204}
\definecolor{steelblue31119180}{RGB}{31,119,180}

\begin{axis}[
width=0.24\textwidth,
height=0.2\textwidth,
scale only axis=true,
tick align=outside,
tick pos=left,
x grid style={darkgray176},
xlabel={$d\vphantom{ / \mathrm{cm}}$},
xmajorgrids,
xmin=-0.0025, xmax=0.0525,
xtick style={color=black},
xticklabel=$\mathclap{\pgfmathprintnumber{\tick}}\vphantom{\degree{}}$,
y grid style={darkgray176},
ymajorgrids,
ymin=-0.05, ymax=1.05,
ytick style={color=black},
change x base,
unit markings=slash space,
x SI prefix=centi,x unit=m,
every x tick scale label/.append style={overlay}
]
\addplot [very thick, steelblue31119180]
table {%
0 0
0.005 0.0392108202010175
0.01 0.188050626628614
0.015 0.376597592753443
0.02 0.549571907184514
0.025 0.690780493857799
0.03 0.797369400670058
0.035 0.873309343591016
0.04 0.92592133018985
0.045 0.952723662985482
0.05 0.969847375604914
};
\addplot [very thick, darkorange25512714]
table {%
0 0
0.005 0.0959796500806552
0.01 0.357798734334285
0.015 0.558506018116392
0.02 0.738615212805559
0.025 0.845700459114034
0.03 0.89421764486909
0.035 0.924184141953096
0.04 0.944099764238739
0.045 0.954274723911155
0.05 0.96165777391736
};
\addplot [very thick, forestgreen4416044]
table {%
0 0
0.005 0.0259337386772552
0.01 0.130537287504653
0.015 0.313376349423005
0.02 0.529035860528602
0.025 0.701203623278322
0.03 0.805062662861397
0.035 0.871634197791289
0.04 0.909480084377714
0.045 0.927906688174712
0.05 0.943231170120362
};
\end{axis}%
\end{tikzpicture}%
%
\begin{tikzpicture}

\definecolor{crimson2143940}{RGB}{214,39,40}
\definecolor{darkgray176}{RGB}{176,176,176}
\definecolor{darkorange25512714}{RGB}{255,127,14}
\definecolor{forestgreen4416044}{RGB}{44,160,44}
\definecolor{lightgray204}{RGB}{204,204,204}
\definecolor{steelblue31119180}{RGB}{31,119,180}

\begin{axis}[
width=0.24\textwidth,
height=0.2\textwidth,
scale only axis=true,
tick align=outside,
tick pos=left,
x grid style={darkgray176},
xlabel={$\delta\vphantom{ / \mathrm{cm}}$},
xticklabel=$\mathclap{\pgfmathprintnumber{\tick}\degree{}}$,
xmajorgrids,
xmin=-1, xmax=21,
xtick style={color=black},
y grid style={darkgray176},
ymajorgrids,
ymin=-0.05, ymax=1.05,
ytick style={color=black}
]
\addplot [very thick, steelblue31119180]
table {%
0 0
1 0.00651445588782727
2 0.0256855689291475
3 0.0576374239980146
4 0.108698349671175
5 0.168879513587294
6 0.232721181288001
7 0.286139719568185
8 0.335835711626753
9 0.381064648219382
10 0.425983372626877
11 0.467117508375729
12 0.504529097902966
13 0.534681722298052
14 0.561422012656657
15 0.585742647971212
16 0.610497580344956
17 0.636493361459238
18 0.656533068618935
19 0.678185879141333
20 0.700024816974811
};
\addplot [very thick, darkorange25512714]
table {%
0 0
1 0.0258716962402283
2 0.0954212681474128
3 0.176448690904579
4 0.264300781734707
5 0.346569053232411
6 0.421826529346073
7 0.489204615957315
8 0.546593870207222
9 0.595607395458494
10 0.632770815237623
11 0.665777391735947
12 0.694068743020226
13 0.718513463208835
14 0.738118873309344
15 0.760578235513091
16 0.779501178806304
17 0.794639533440874
18 0.808164784712744
19 0.821255738925425
20 0.833850353641891
};
\addplot [very thick, forestgreen4416044]
table {%
0 0
1 0.0332547462464326
2 0.123402407246557
3 0.228874550192332
4 0.330065764983249
5 0.420957935227696
6 0.49274103486785
7 0.538652438267775
8 0.570169996277454
9 0.592133018984986
10 0.612979277826033
11 0.62979277826033
12 0.645241345080035
13 0.661310336270009
14 0.67682094552674
15 0.698908053108326
16 0.720933118252885
17 0.738677255242586
18 0.75393969475121
19 0.76343218761633
20 0.772614468296315
};
\end{axis}%
\end{tikzpicture}%
%
\begin{tikzpicture}

\definecolor{crimson2143940}{RGB}{214,39,40}
\definecolor{darkgray176}{RGB}{176,176,176}
\definecolor{darkorange25512714}{RGB}{255,127,14}
\definecolor{forestgreen4416044}{RGB}{44,160,44}
\definecolor{lightgray204}{RGB}{204,204,204}
\definecolor{steelblue31119180}{RGB}{31,119,180}

\begin{axis}[
width=0.24\textwidth,
height=0.2\textwidth,
scale only axis=true,
tick align=outside,
tick pos=left,
x grid style={darkgray176},
xticklabel=$\mathclap{\pgfmathprintnumber{\tick}}\vphantom{\degree{}}$,
xlabel={$F_{1\mathrm{cm}}\vphantom{ / \mathrm{cm}}$},
xmajorgrids,
xmin=-0.05, xmax=1.05,
xtick style={color=black},
y grid style={darkgray176},
ymajorgrids,
ymin=-0.05, ymax=1.05,
ytick style={color=black}
]
\addplot [very thick, steelblue31119180]
table {%
0 1
0.05 0.943355254994416
0.1 0.912147909169872
0.15 0.881188733093436
0.2 0.827025685568929
0.25 0.744199032137982
0.3 0.6481573396203
0.35 0.541258220622906
0.4 0.426914009182281
0.4 0.426914009182281
0.45 0.334160565827026
0.5 0.254436034247425
0.55 0.154237498448939
0.6 0.0664474500558382
0.65 0.0264921206104976
0.7 0.00570790420647723
0.75 0.000930636555403896
0.8 0
0.85 0
0.9 0
0.95 0
1 0
};
\addplot [very thick, darkorange25512714]
table {%
0 1
0.05 0.991189973942176
0.1 0.984117136121107
0.15 0.976547958803822
0.2 0.962154113413575
0.25 0.939260454150639
0.3 0.919965256235265
0.35 0.890308971336394
0.4 0.848678496091327
0.4 0.848678496091327
0.45 0.804566323365182
0.5 0.756979774165529
0.55 0.703809405633453
0.6 0.649894527857054
0.65 0.595235140836332
0.7 0.526926417669686
0.75 0.460044670554659
0.8 0.398250403275841
0.85 0.329321255738925
0.9 0.24041444347934
0.95 0.112731108077925
1 0
};
\addplot [very thick, forestgreen4416044
]
table {%
0 1
0.05 0.973942176448691
0.1 0.964139471398436
0.15 0.953282044918724
0.2 0.938391860032262
0.25 0.917297431443107
0.3 0.882367539396948
0.35 0.832919717086487
0.4 0.750403275840675
0.4 0.750403275840675
0.45 0.647102618190843
0.5 0.55168135004343
0.55 0.449869710882243
0.6 0.356744012904827
0.65 0.278074202754684
0.7 0.208462588410473
0.75 0.156222856433801
0.8 0.114220126566572
0.85 0.0797865740166274
0.9 0.053108326095049
0.95 0.0289117756545477
1 0
};
\end{axis}%
\end{tikzpicture}%
%
        \caption{REAL275}
    \end{subfigure}
    \begin{subfigure}{\textwidth}
        \centering
\begin{tikzpicture}
\definecolor{crimson2143940}{RGB}{214,39,40}
\definecolor{darkgray176}{RGB}{176,176,176}
\definecolor{darkorange25512714}{RGB}{255,127,14}
\definecolor{forestgreen4416044}{RGB}{44,160,44}
\definecolor{lightgray204}{RGB}{204,204,204}
\definecolor{steelblue31119180}{RGB}{31,119,180}

\begin{axis}[
width=0.24\textwidth,
height=0.2\textwidth,
scale only axis=true,
tick align=outside,
tick pos=left,
x grid style={darkgray176},
xlabel={$d\vphantom{ / \mathrm{cm}}$},
xmajorgrids,
xmin=-0.0025, xmax=0.0525,
xtick style={color=black},
xticklabel=$\mathclap{\pgfmathprintnumber{\tick}}\vphantom{\degree{}}$,
y grid style={darkgray176},
ymajorgrids,
ymin=-0.05, ymax=1.05,
ytick style={color=black},
change x base,
unit markings=slash space,
x SI prefix=centi,x unit=m,
every x tick scale label/.append style={overlay}
]
\addplot [very thick, steelblue31119180]
table {%
0 0
0.005 0
0.01 0.0266666666666667
0.015 0.146666666666667
0.02 0.28
0.025 0.413333333333333
0.03 0.573333333333333
0.035 0.746666666666667
0.04 0.853333333333333
0.045 0.893333333333333
0.05 0.933333333333333
};\label{cass}
\addplot [very thick, darkorange25512714]
table {%
0 0
0.005 0.0133333333333333
0.01 0.0933333333333333
0.015 0.28
0.02 0.413333333333333
0.025 0.56
0.03 0.666666666666667
0.035 0.693333333333333
0.04 0.8
0.045 0.853333333333333
0.05 0.866666666666667
};\label{spd}
\addplot [very thick, forestgreen4416044]
table {%
0 0
0.005 0.0266666666666667
0.01 0.16
0.015 0.346666666666667
0.02 0.586666666666667
0.025 0.626666666666667
0.03 0.693333333333333
0.035 0.773333333333333
0.04 0.813333333333333
0.045 0.84
0.05 0.866666666666667
};\label{asmnet}
\end{axis}%
\end{tikzpicture}%
%
\begin{tikzpicture}

\definecolor{crimson2143940}{RGB}{214,39,40}
\definecolor{darkgray176}{RGB}{176,176,176}
\definecolor{darkorange25512714}{RGB}{255,127,14}
\definecolor{forestgreen4416044}{RGB}{44,160,44}
\definecolor{lightgray204}{RGB}{204,204,204}
\definecolor{steelblue31119180}{RGB}{31,119,180}

\begin{axis}[
width=0.24\textwidth,
height=0.2\textwidth,
scale only axis=true,
tick align=outside,
tick pos=left,
x grid style={darkgray176},
xlabel={$\delta\vphantom{ / \mathrm{cm}}$},
xticklabel=$\mathclap{\pgfmathprintnumber{\tick}\degree{}}$,
xmajorgrids,
xmin=-1, xmax=21,
xtick style={color=black},
y grid style={darkgray176},
ymajorgrids,
ymin=-0.05, ymax=1.05,
ytick style={color=black}
]
\addplot [very thick, steelblue31119180]
table {%
0 0
1 0
2 0
3 0
4 0
5 0
6 0
7 0.0133333333333333
8 0.0266666666666667
9 0.0266666666666667
10 0.0266666666666667
11 0.0266666666666667
12 0.0533333333333333
13 0.0666666666666667
14 0.0666666666666667
15 0.0666666666666667
16 0.0666666666666667
17 0.0666666666666667
18 0.0666666666666667
19 0.08
20 0.106666666666667
};
\addplot [very thick, darkorange25512714]
table {%
0 0
1 0
2 0.0266666666666667
3 0.0266666666666667
4 0.0533333333333333
5 0.0666666666666667
6 0.0933333333333333
7 0.133333333333333
8 0.173333333333333
9 0.186666666666667
10 0.2
11 0.213333333333333
12 0.24
13 0.266666666666667
14 0.266666666666667
15 0.266666666666667
16 0.293333333333333
17 0.293333333333333
18 0.306666666666667
19 0.306666666666667
20 0.306666666666667
};
\addplot [very thick, forestgreen4416044]
table {%
0 0
1 0.0266666666666667
2 0.0933333333333333
3 0.12
4 0.2
5 0.24
6 0.266666666666667
7 0.333333333333333
8 0.346666666666667
9 0.386666666666667
10 0.386666666666667
11 0.386666666666667
12 0.4
13 0.426666666666667
14 0.426666666666667
15 0.453333333333333
16 0.466666666666667
17 0.466666666666667
18 0.466666666666667
19 0.546666666666667
20 0.546666666666667
};
\end{axis}%
\end{tikzpicture}%
%
\begin{tikzpicture}

\definecolor{crimson2143940}{RGB}{214,39,40}
\definecolor{darkgray176}{RGB}{176,176,176}
\definecolor{darkorange25512714}{RGB}{255,127,14}
\definecolor{forestgreen4416044}{RGB}{44,160,44}
\definecolor{lightgray204}{RGB}{204,204,204}
\definecolor{steelblue31119180}{RGB}{31,119,180}

\begin{axis}[
width=0.24\textwidth,
height=0.2\textwidth,
scale only axis=true,
tick align=outside,
tick pos=left,
x grid style={darkgray176},
xticklabel=$\mathclap{\pgfmathprintnumber{\tick}}\vphantom{\degree{}}$,
xlabel={$F_{1\mathrm{cm}}\vphantom{ / \mathrm{cm}}$},
xmajorgrids,
xmin=-0.05, xmax=1.05,
xtick style={color=black},
y grid style={darkgray176},
ymajorgrids,
ymin=-0.05, ymax=1.05,
ytick style={color=black}
]
\addplot [very thick, steelblue31119180]
table {%
0 1
0.05 0.773333333333333
0.1 0.693333333333333
0.15 0.626666666666667
0.2 0.546666666666667
0.25 0.48
0.3 0.373333333333333
0.35 0.28
0.4 0.213333333333333
0.4 0.213333333333333
0.45 0.186666666666667
0.5 0.12
0.55 0.0533333333333333
0.6 0.0133333333333333
0.65 0.0133333333333333
0.7 0
0.75 0
0.8 0
0.85 0
0.9 0
0.95 0
1 0
};
\addplot [very thick, darkorange25512714]
table {%
0 1
0.05 0.986666666666667
0.1 0.986666666666667
0.15 0.906666666666667
0.2 0.893333333333333
0.25 0.84
0.3 0.733333333333333
0.35 0.64
0.4 0.573333333333333
0.4 0.573333333333333
0.45 0.533333333333333
0.5 0.506666666666667
0.55 0.493333333333333
0.6 0.44
0.65 0.36
0.7 0.226666666666667
0.75 0.186666666666667
0.8 0.146666666666667
0.85 0.0666666666666667
0.9 0.04
0.95 0.0266666666666667
1 0
};
\addplot [very thick, forestgreen4416044
]
table {%
0 1
0.05 0.986666666666667
0.1 0.973333333333333
0.15 0.933333333333333
0.2 0.933333333333333
0.25 0.92
0.3 0.826666666666667
0.35 0.813333333333333
0.4 0.746666666666667
0.4 0.746666666666667
0.45 0.733333333333333
0.5 0.666666666666667
0.55 0.533333333333333
0.6 0.453333333333333
0.65 0.386666666666667
0.7 0.226666666666667
0.75 0.186666666666667
0.8 0.106666666666667
0.85 0.04
0.9 0.0133333333333333
0.95 0
1 0
};
\end{axis}%
\end{tikzpicture}%
%
        \caption{REDWOOD75}
    \end{subfigure}
    \caption{Detailed precision results for CASS (\ref{cass}), SPD (\ref{spd}) and ASM-Net (\ref{asmnet}) for varying thresholds of position, orientation and F-score thresholds.}
    \label{fig:varying_thresholds}
\end{figure}

To get further insights into the estimation quality of the methods we show detailed results for varying thresholds in Fig.\ \ref{fig:varying_thresholds}. It can be seen that the difference between the two datasets is most pronounced for orientation-based thresholding. This confirms the issue of constrained orientations discussed in Section \ref{section:datasets} (see Fig.\ \ref{fig:orientation_dist}). ASM-Net, which was trained on synthetic data (exact distribution unspecified, but likely less constrained than the CAMERA and REAL datasets) performs best in this metric.

In Table \ref{tab:categories} we further report per-category precision with the more lenient pose and shape estimation thresholds $\delta=10\degree, d=\SI{2}{cm}, F_{1\mathrm{cm}}=0.6$. Note that CASS' reconstructions are very sparse and noisy, and therefore rarely reach $F_{1\mathrm{cm}}>0.6$ (see also Fig.\ \ref{fig:varying_thresholds}). On REAL275, all methods fail at the camera category which contains significantly more shape variation than the other categories. Note that despite these relatively lenient thresholds, all methods fail to sufficiently recover pose and shape for most of the REDWOOD75 samples.

\begin{table}[htb]
    \centering
    \caption{Precision for different categories with $\delta=10^{\circ},d=\SI{2}{cm},F_{1\mathrm{cm}}=0.6$.}\label{tab:categories}
    \scriptsize
    \begin{tabular}{@{}lR{1cm}R{1cm}R{1cm}R{1cm}R{1cm}R{1cm}cR{1cm}R{1cm}R{1cm}@{}}
        \toprule
         & \multicolumn{6}{c}{REAL275} & \phantom{a} & \multicolumn{3}{c}{REDWOOD75} \\
         \cmidrule{2-7} \cmidrule{9-11}
         & Bottle & Bowl & Camera & Can & Laptop & Mug & & Bottle & Bowl & Mug \\
         \midrule
         CASS & 0.002 & 0.093 & 0.001 & 0.030 & 0.0 & 0.068 & & 0.000 & 0.000 & 0.000 \\
         SPD & \textbf{0.610} & \textbf{0.892} & \textbf{0.052} & \textbf{0.863} & \textbf{0.218} & \textbf{0.246} & & \textbf{0.320} & 0.160 & \textbf{0.040} \\
         ASM-Net & 0.167 & 0.137 & 0.023& 0.587 & 0.133 & 0.229 & & 0.160 & \textbf{0.360} & 0.000 \\
         \bottomrule

    \end{tabular}
\end{table}

\section{Limitations}\label{sec:limitations}

\paragraph{Comparability} The results from Section \ref{sec:experiments} suggest that training on synthetic data currently generalizes better to unconstrained orientations. This is expected, since synthetic data allows accurate, unconstrained generation of training data. This opens the question, whether CASS or SPD would generalize better or worse when trained on the same synthetic data as ASM-Net. Since the training code for CASS and ASM-Net was not published it is difficult to perform further comparisons. In general, since methods currently vary significant parts of training datasets, architecture, pose parameterization and losses, it is difficult to assess the individual contributions of single changes.

\paragraph{Multimodal distributions} Unconstrained pose estimation introduces significant difficulties to the task, which were hidden due to the constraints present in the CAMERA and REAL datasets. Consider, for example, the bottom-left mug in Fig.\ \ref{fig:redwood_examples}. From the given view, it is difficult to tell which way the opening of the mug faces. Another example are cans, which are geometry-wise nearly symmetric. To our knowledge, there currently exists no method that allows categorical multimodal pose or shape inference. A possible way of evaluating such methods would be to allow methods to generate $N$ hypotheses. Precision could then be computed for the best and worst hypotheses. A strong method would generate the same hypothesis $N$ times if there is no ambiguity. If there is ambiguity, the correct hypothesis would still be contained in the set of hypotheses with a high likelihood.

\paragraph{Dataset size} The REDWOOD75 dataset is limited in size, but the results suggest a clear lack of generalization capability of current approaches. This shows the need for larger datasets for unconstrained pose and shape estimation. It is an open question how such a dataset could be collected in the most efficient way.









\section{Conclusion and Outlook}
In this work we have discussed the limitations of the current evaluation protocol prevalent in the field of categorical pose and shape estimation. In particular, existing datasets only contain a heavily constrained set of orientations which simplifies the problem by removing pose and shape ambiguities. Furthermore, existing evaluation metrics are suboptimal and unnecessarily difficult to interpret. To alleviate these problems, we propose a new set of metrics applicable to both the established REAL275 and our proposed REDWOOD75 dataset, which contains a large variety of orientations. We apply our evaluation protocol to three methods and confirm limited generalization capability as suggested by the constrained orientations in their training data.

Our experiments suggest that there is a need for larger, high-quality datasets for unconstrained pose and shape estimation as well as for methods that can handle unconstrained orientations and the resulting pose ambiguities in a principled way.


%
%
\bibliographystyle{splncs03}
\bibliography{bib.bib}

\end{document}